\documentclass{article}

    \PassOptionsToPackage{numbers, compress}{natbib}

\usepackage[preprint]{neurips_2026}

\usepackage[utf8]{inputenc}
\usepackage[T1]{fontenc}
\usepackage[hidelinks]{hyperref}
\usepackage{url}
\usepackage{booktabs}
\usepackage{amsfonts}
\usepackage{amsmath}
\usepackage{amssymb}
\usepackage{nicefrac}
\usepackage{microtype}
\usepackage{xcolor}
\usepackage{algorithm}
\usepackage{algpseudocode}
\usepackage{graphicx}
\usepackage{xspace}
\usepackage{multirow}
\usepackage{wrapfig}
\usepackage{colortbl}
\usepackage{subcaption}

\usepackage{tcolorbox}
\tcbuselibrary{breakable, skins}
\usepackage{mdframed}
\usepackage{listings}
\definecolor{csProblemBg}{RGB}{255, 250, 240}
\definecolor{csVanillaBg}{RGB}{235, 240, 255}
\definecolor{csExpBg}{RGB}{235, 250, 235}
\definecolor{csVanillaBlue}{RGB}{40, 80, 180}
\definecolor{csExpGreen}{RGB}{30, 130, 60}
\lstdefinestyle{cscode}{
  basicstyle=\tiny\ttfamily,
  breaklines=true,
  breakatwhitespace=false,
  keepspaces=true,
  showspaces=false,
  showstringspaces=false,
  tabsize=2,
  frame=none,
  backgroundcolor=\color{gray!8},
}

\title{ExpThink: Experience-Guided Reinforcement Learning for Adaptive Chain-of-Thought Compression}

\newcommand{\ourMethod}{\textbf{ExpThink}\xspace}

\author{%
  Tingcheng Bian\textsuperscript{1,2}\thanks{Equal contribution.} \\
  \texttt{bincarnegie@gmail.com}
  \And
  Yuzhe Zhang\textsuperscript{1,3}\footnotemark[1] \\
  \texttt{zhangyuzhe02@baidu.com}
  \And
  Jing Jin\textsuperscript{4}\footnotemark[1] \\
  \texttt{jingjin0007@gmail.com}
  \AND
  Jinchang Luo\textsuperscript{1} \\
  \texttt{luojinchang@baidu.com}
  \And
  Mingquan Cheng\textsuperscript{1} \\
  \texttt{chengmingquan@baidu.com}
  \And
  Haiwei Wang\textsuperscript{1}\thanks{Corresponding authors.} \\
  \texttt{wanghaiwei@baidu.com}
  \And
  Wenyuan Jiang\textsuperscript{5}\footnotemark[2] \\
  \texttt{wenyjiang@ethz.ch}
  \And
  Miaohui Wang\textsuperscript{2} \\
  \texttt{wang.miaohui@gmail.com}
  \AND
  \normalsize\mdseries
  \textsuperscript{1}Baidu Inc. \quad
  \textsuperscript{2}Shenzhen University \quad
  \textsuperscript{3}Peking University \\[2pt]
  \textsuperscript{4}Tsinghua University \quad
  \textsuperscript{5}D-INFK, ETH Z\"{u}rich
}

\begin{document}

\maketitle

\begin{abstract}

Large reasoning models (LRMs) achieve strong performance via extended
chain-of-thought (CoT) reasoning, yet suffer from excessive token consumption
and high inference latency. Existing reinforcement learning (RL) approaches
for CoT compression rely on uniform, static length penalties that neglect
model capability dynamics and problem-level difficulty variation.
We propose \ourMethod, an RL framework that addresses both dimensions through
two complementary mechanisms.
First, \emph{experience-guided reward shaping} tracks the shortest correct
solution found so far for each problem and applies a three-tier reward:
full credit for concise correct responses, discounted credit for verbose
correct ones, and zero for incorrect ones. The threshold tightens
automatically with model improvement, forming a self-evolving curriculum
that requires no manual scheduling.
Second, \emph{difficulty-adaptive advantage} replaces standard deviation
normalization with correct-count normalization, yielding
monotonically difficulty-scaled gradients that amplify learning on hard
problems to preserve accuracy while suppressing gradients on easy ones to
encourage brevity.
Together, these mechanisms enforce an accuracy-first, compression-second
training objective.
Experiments on multiple mathematical reasoning benchmarks demonstrate that
\ourMethod reduces average response length by up to 77\% while simultaneously
improving accuracy, achieving up to $3\times$ higher accuracy-efficiency ratio (accuracy divided by average token count) than the vanilla baseline and outperforming existing RL-based compression
methods on both metrics.

\end{abstract}

\section{Introduction}
\label{sec:intro}

Large Reasoning Models (LRMs) such as OpenAI-o1, DeepSeek-R1, and Qwen3~\cite{jaech2024openai,guo2025deepseek,yang2025qwen3} have shown that extended Chain-of-Thought (CoT) sequences substantially improve complex reasoning~\cite{wei2022chain}. However, these models produce lengthy reasoning traces regardless of problem difficulty, incurring redundant computation, high latency, and verbose outputs~\cite{fu2022complexity,fu2024efficiently,fu2025reasoning}. We quantify this inefficiency via \emph{intelligence per token} (IPT), defined as the ratio of correctness to generation length, and find that current LRMs achieve strong accuracy at the cost of poor IPT.

\begin{figure*}[!t]
    \centering
    \includegraphics[width=0.95\linewidth]{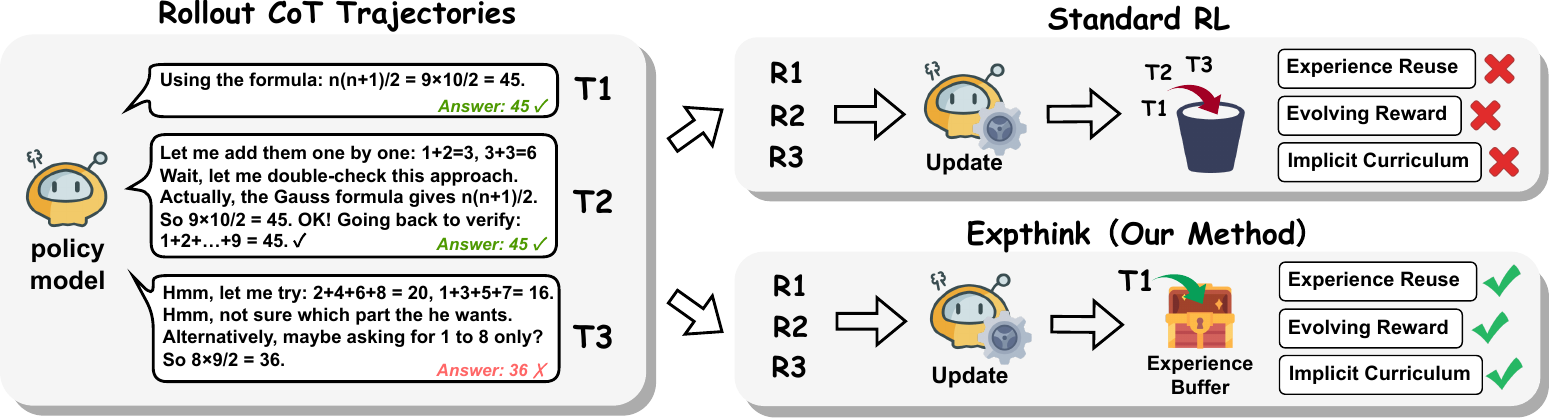}
    \caption{
    \textbf{Top:} Standard RL treats each epoch independently, discarding all trajectory information after each update. \textbf{Bottom:} \ourMethod{} accumulates successful trajectories into an experience buffer, enabling a self-evolving compression curriculum that tightens automatically as the policy improves.
}
    \label{fig:introduction}
\end{figure*}

Inference-time strategies such as adaptive routing and early exits reduce output length but leave the model's inherent verbosity unchanged~\cite{zeyu2025large,huang2025efficient,zhang2025alphaone}. Supervised fine-tuning on distilled concise traces is prone to distribution shift and depends heavily on demonstration quality~\cite{wu2025efficient,zhang2025dart}. Reinforcement learning (RL) offers a more principled path by jointly rewarding correctness and brevity, and methods such as L1, ThinkPrune, LC-R1, and Laser have demonstrated substantial token reduction with minimal accuracy loss~\cite{aggarwal2025l1,dumitru2025conciserl,hou2025thinkprune,liu2025learn}. Yet all existing RL approaches share a structural limitation (Figure~\ref{fig:introduction}): each epoch is treated as an independent episode, discarding all trajectory information after every update.

This epoch-independent design neglects a basic fact: the policy's capability is not stationary. As training progresses, problems that once required lengthy derivations become solvable in fewer steps, yet fixed length penalties assume a static compression goal throughout. The training process itself already produces exactly the signal needed: each time the model finds a short correct solution, it establishes a tighter per-problem compression bound, tracing a natural coarse-to-fine curriculum. Standard RL discards this signal after every update, never using it to recalibrate the reward.

We propose \ourMethod{} to exploit this signal. The framework maintains an \emph{experience buffer} that records the shortest correct solution per problem across all past epochs, defining an adaptive reward threshold that tightens automatically as the policy improves, without any manual scheduling. Our key contributions are:

\begin{itemize}
    \item We identify a structural limitation shared by existing RL-based CoT compression methods: historical trajectory information accumulated during training is discarded after each update and never used to recalibrate the compression objective as the policy improves.

    \item We propose \ourMethod, an experience-driven RL framework with a self-evolving compression curriculum, an experience-guided three-tier reward, and difficulty-adaptive advantage estimation that jointly optimize accuracy and brevity.

    \item \ourMethod{} reduces average response length by up to 77\% while consistently improving accuracy across all model scales, with stable training dynamics, achieving up to $3\times$ higher IPT than the vanilla baseline and outperforming all prior RL-based compression methods on both metrics.
\end{itemize}

\section{Preliminary Observation}
\label{sec:preliminary}

We begin with an empirical observation that motivates our approach. Using the setup described in Section~\ref{sec:experiments}, we train a reasoning model on DeepScaleR for approximately 55 steps and track two per-prompt statistics: the \emph{batch mean length} $\bar{\ell}$, the mean response length over all prompts at the current step, and the \emph{mean shortest correct (historical)} $\ell^{*}$, the running minimum of correct response length across all past training steps averaged over prompts.

\begin{wrapfigure}{r}{0.52\textwidth}
  \vspace{-8pt}
  \centering
  \includegraphics[width=0.42\textwidth]{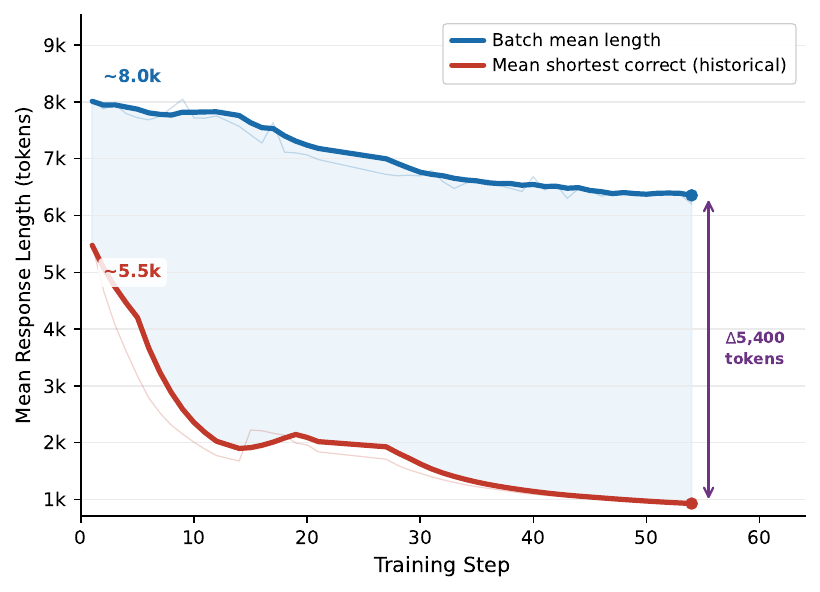}
  \caption{Response length dynamics during RL training on DeepScaleR.}
  \label{fig:preliminary}
  \vspace{-10pt}
\end{wrapfigure}

Figure~\ref{fig:preliminary} reveals a striking divergence between these two curves over the first ${\sim}$55 training steps. The blue curve (batch mean length) starts at approximately 8{,}000 tokens and decreases only modestly to around 6{,}300 tokens by the end of training. 

In contrast, the red curve (mean shortest correct, historical) begins at approximately 5{,}500 tokens and drops sharply to around 1{,}000 tokens, demonstrating that the model rapidly learns to solve problems far more concisely.

By step~55, the gap between the two curves widens to $\Delta$5{,}400 tokens: the model \emph{can} produce short correct solutions, yet its average rollout length remains far above this self-demonstrated lower bound.
This exposes a structural inefficiency: the short correct trajectories already produced during training are discarded after each update and never used to recalibrate the length signal.
This observation motivates a simple principle: \emph{reuse the shortest correct trajectories from past rollouts as an adaptive, per-problem compression target}.
We formalize this idea in the next section.

\section{ExpThink}
\label{sec:method}

The key insight driving \ourMethod is that standard RL training discards two
valuable signals after every update: evidence that certain problems \emph{can}
be solved correctly, which establishes a per-problem achievability bound on
response length, and evidence that the LRM \emph{already solves} many problems
consistently, which identifies where gradient budget is wasted on correctness
and should shift to brevity.
\ourMethod captures both signals through a minimal modification to
DAPO~\cite{yu2025dapo}: a running experience buffer that converts the first
signal into an adaptive reward threshold, and a count-based advantage normalizer
that converts the second into difficulty-weighted gradients.

Throughout this section we assume a training distribution $\mathcal{D}$ of
queries $q$ with ground-truth answers $y^\ast$.
For each $q$, the old policy samples a group of $G$ rollouts
$\{o_i\}_{i=1}^{G}$, where $o_i$ has token length $|o_i|$ and produces a
final answer $y_i$.
We write $|\mathcal{C}_q| = \sum_{i=1}^{G}\mathbb{I}[y_i = y^\ast]$ for the
number of correct rollouts in the group.

\subsection{Experience-Guided Reward Shaping}
\label{sec:exp_compression}

A binary correctness reward treats every correct rollout identically and therefore gives the policy no reason to prefer shorter derivations over longer ones. Length penalties can address this, but the penalty is typically static: it is fixed ahead of time and remains unchanged as the LRM improves, even though the same problem might admit progressively shorter solutions over the course of training. We observe that every correct rollout the LRM produces is already direct evidence of an achievable length bound on that problem, and that these bounds monotonically tighten as the policy improves. \ourMethod\ uses these running bounds as the reward signal.

Concretely, we maintain a global dictionary $\mathcal{E}$ that stores, for every training prompt $q$, the shortest correct trajectory length ever observed. Writing $\ell^{*,(t)}_q$ for this value at step $t$, the update after each batch is
\begin{equation}
    \ell^{*,(t)}_q \;=\; \min\!\left(\, \ell^{*,(t-1)}_q,\;\; \min_{\,i\,:\,y_i = y^\ast\,}\,|o_i| \,\right),
    \label{eq:buffer_update}
\end{equation}
where the inner minimum is $+\infty$ whenever no rollout in the current group is correct, leaving the buffer unchanged. Prompts that have never yielded a correct trajectory are initialized with $\ell^{*,(0)}_q = L_{\max}$, the training context budget. Because \eqref{eq:buffer_update} takes a running minimum, $\ell^{*}_q$ is non-increasing throughout training and traces an implicit, per-prompt curriculum whose target tightens exactly as the LRM discovers shorter solutions, without any manually scheduled annealing.

The buffer defines a per-problem achievability bound, but a binary correct/incorrect
reward cannot exploit it: declaring a verbose-correct response ``wrong'' would strip the
policy of its only positive signal on hard problems where short solutions have not yet
been discovered. Instead, we define a three-tier reward that preserves positive signal
for every correct response while imposing graded compression pressure:
\begin{equation}
    r_i^{\text{exp}} \;=\; \begin{cases}
        1,              & y_i = y^\ast \;\text{and}\; |o_i| \leq \ell^*_q(1+\alpha), \\[2pt]
        r_{\text{pen}}, & y_i = y^\ast \;\text{and}\; |o_i| > \ell^*_q(1+\alpha), \\[2pt]
        0,              & y_i \neq y^\ast.
    \end{cases}
    \label{eq:three_tier}
\end{equation}
The tolerance $\alpha \in [0,+\infty)$ provides a soft margin around the historical minimum, and $r_{\text{pen}} \in [0,1)$ assigns correct but verbose rollouts a discounted reward. The discount distinguishes concise solutions from verbose ones without outright punishing correctness, so the policy still receives positive signal whenever it gets the answer right.

\begin{figure}[!t]
    \centering
    \includegraphics[width=\linewidth]{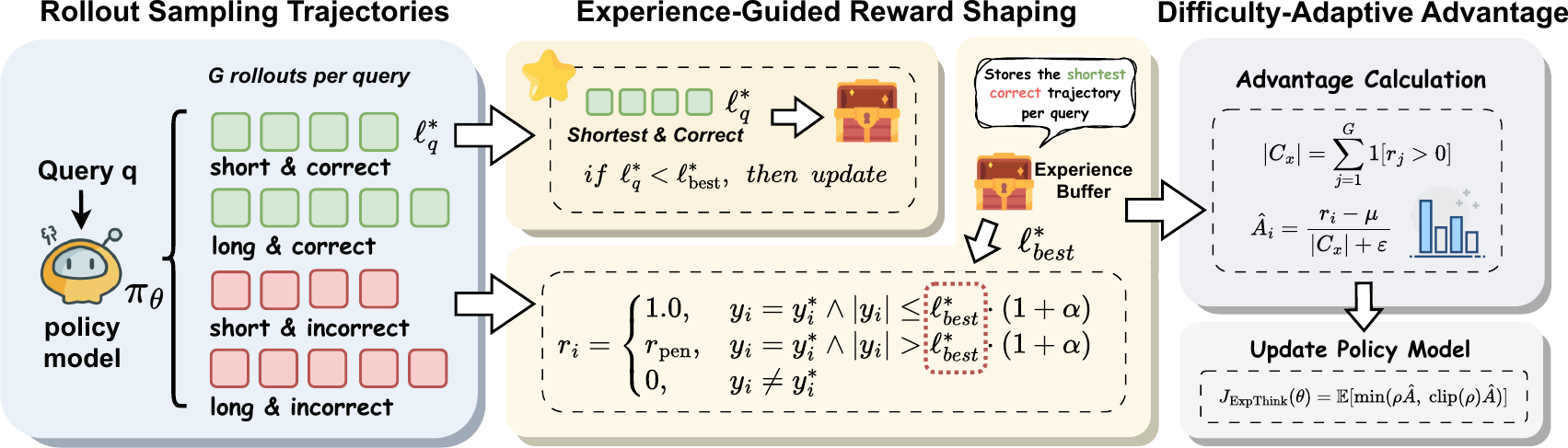}
    \caption{Overview of the ExpThink framework. For each query, the policy samples a group of rollouts, from which an experience buffer extracts the shortest correct trajectory seen so far; this running minimum is used to shape a pass-rate gated, three-tier reward, which is in turn combined with a count-based advantage estimator to produce difficulty-adaptive gradients.}
    \label{fig:overview}
\end{figure}

\subsection{Difficulty-Adaptive Advantage}
\label{sec:DA_norm}

Prior work has proposed reshaping the reward to reflect problem difficulty~\cite{liu2025learn,liu2025dler}, yet intervening at the reward level is insufficient: the per-group standard deviation $\sigma$ used in GRPO is a concave function of $|\mathcal{C}_q|$, peaking when half the rollouts are correct and vanishing at both extremes. A prompt the policy almost never solves and one it almost always solves therefore receive nearly identical normalization, and whatever difficulty signal was injected at the reward level is silently cancelled before reaching the gradient. We resolve this by replacing $\sigma$ with $|\mathcal{C}_q|$ directly:
\begin{equation}
    \hat{A}_i \;=\; \frac{r_i - \mu}{|\mathcal{C}_q| + \epsilon}, \qquad
    \mu \;=\; \frac{1}{G}\sum_{j=1}^{G} r_j,
    \label{eq:cc_adv}
\end{equation}
where $\epsilon > 0$ and the denominator is clamped to $1$ when $|\mathcal{C}_q| = 0$. Because $|\mathcal{C}_q|$ counts exactly how many rollouts the policy solved correctly, gradient magnitude is now strictly decreasing in $|\mathcal{C}_q|$, preserving difficulty awareness all the way to the parameter update. This substitution also reveals a clean structural property: the advantage gap between a short-correct and a long-correct rollout in the same group equals $(1-r_{\mathrm{pen}})/(|\mathcal{C}_q|+\epsilon)$, which is the selective pressure the gradient exerts on length. On hard prompts this gap is a minor fraction of a large overall advantage, so the update drives correctness; on easy prompts the overall advantage shrinks and this gap becomes the dominant signal, so the update drives brevity. We term this the \emph{accuracy-first, compression-second} learning dynamic: the LRM is pushed toward conciseness only where correctness is already reliable, with no manual scheduling required. In combination with the self-evolving threshold from Eq.~\eqref{eq:buffer_update}, the two mechanisms jointly induce an implicit curriculum along both the accuracy and length axes.

Figure~\ref{fig:overview} illustrates the complete per-step data flow, highlighting how
the experience buffer and the advantage normalizer interact within the DAPO training loop.

\subsection{Training Objective}
\label{sec:objective}

The shaped reward $r_i$ and the reweighted advantage $\hat{A}_i$ replace their counterparts in the standard DAPO~\cite{yu2025dapo} objective. Broadcasting $\hat{A}_i$ across all tokens of $o_i$ to obtain $\hat{A}_{i,t}$, and writing $\rho_{i,t}(\theta) = \pi_\theta(o_{i,t}\mid q, o_{i,<t}) / \pi_{\theta_{\text{old}}}(o_{i,t}\mid q, o_{i,<t})$ for the token-level importance ratio, \ourMethod\ optimizes
\begin{equation}
\begin{aligned}
  \mathcal{J}(\theta) \;=\;
  &\mathbb{E}_{(q,\,y^\ast)\sim\mathcal{D},\,\{o_i\}_{i=1}^{G}\sim\pi_{\theta_{\text{old}}}(\cdot\mid q)}\\
  &\left[\,
    \frac{1}{\sum_{i=1}^{G}|o_i|}
    \sum_{i=1}^{G}\sum_{t=1}^{|o_i|}
    \min\!\left(
      \rho_{i,t}(\theta)\,\hat{A}_{i,t},\;
      \operatorname{clip}\!\left(
        \rho_{i,t}(\theta),\,
        1 - \varepsilon_{\text{low}},\,
        1 + \varepsilon_{\text{high}}
      \right)\hat{A}_{i,t}
    \right)\,
  \right].
\end{aligned}
\label{eq:dapo_obj}
\end{equation}
The asymmetric clipping thresholds $\varepsilon_{\text{low}} < \varepsilon_{\text{high}}$ are inherited from DAPO and permit larger updates in the favorable direction while retaining a conservative lower bound on the unfavorable side.

\section{Experiments}
\label{sec:experiments}

\subsection{Setup}
\label{sec:experiments:setup}

\paragraph{Models.}
We evaluate \ourMethod on DeepSeek-R1-Distill-Qwen-1.5B, DeepSeek-R1-Distill-Qwen-7B and Qwen3-8B~\citep{guo2025deepseek, yang2025qwen3} as backbone models.

\paragraph{Training.}
Training is conducted on the DeepScaleR dataset~\citep{luo2025deepscaler}, which contains approximately 40K mathematical problems with calibrated difficulty levels. We set $\alpha=0.1$ and $r_{\text{pen}}=0.5$ as default; full hyperparameters are provided in Appendix~\ref{app:training_params}.

\paragraph{Evaluation benchmarks.}
We evaluate on five standard mathematical reasoning benchmarks: AIME24~\citep{codeforcesamerican}, AMC23~\citep{amc23}, MATH-500~\citep{lightman2023let}, Minerva Math~\citep{lewkowycz2022solving}, and OlympiadBench~\citep{he2024olympiadbench}. To further assess out-of-domain generalization, we additionally evaluate on LiveCodeBench~\citep{jain2024livecodebench} for code reasoning, GPQA-Diamond~\citep{rein2024gpqa} for graduate-level scientific question answering, and MMLU~\citep{hendrycks2020measuring} for broad multitask language understanding. Detailed descriptions of all benchmarks are provided in Appendix~\ref{app:datasets}.

\paragraph{Evaluation protocol.}
We report Pass@1 accuracy, average response token length, and Intelligence per Token (IPT), defined as accuracy divided by response length in kilotokens. To quantify efficiency gains, we additionally report $\Delta$Acc and $\Delta$Token as the average relative change (\%) across all benchmarks with respect to the Vanilla baseline.
Evaluation runs are replicated 16 times on AIME24, 8 times on AMC23, and once on the remaining benchmarks, which have larger test sets. Context length and sampling hyperparameters follow DeepSeek-R1~\citep{guo2025deepseek}.

\paragraph{Baselines.}
We compare against five representative RL-based methods spanning aggressive and conservative length-control strategies: LC-R1~\citep{cheng2025optimizing}, Laser~\citep{liu2025learn}, AutoThink~\citep{tu2025learning}, AdaptThink~\citep{zhang2025adaptthink}, and JET~\citep{han2025your}. Detailed descriptions of each baseline are provided in Appendix~\ref{app:baselines}.

\subsection{Main Results}

\begin{figure}[!t]
    \centering
    \includegraphics[width=0.95\linewidth]{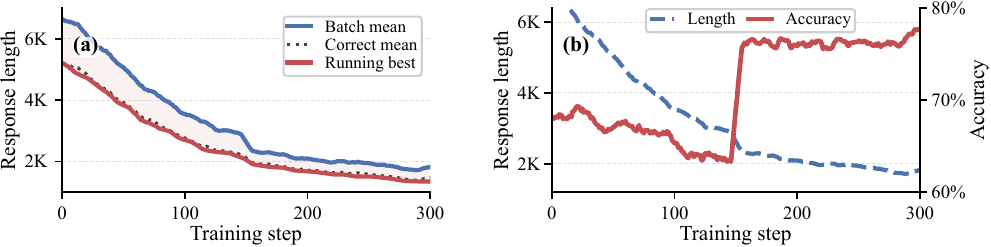}
    \caption{
    Experience buffer dynamics during \ourMethod training.
    (a) Step-level response length trajectories; the running best-correct curve
    marks the tightening length target maintained by the experience buffer.
    (b) Average batch length and accuracy across training.
}
    \label{fig:buffer_dynamics}
\end{figure}

Table~\ref{tab:main_results} and Figure~\ref{fig:buffer_dynamics} reveal three consistent patterns across all model scales.

\textbf{(1) Simultaneous accuracy and compression gains.}
\ourMethod{} is the only method that simultaneously improves accuracy \emph{and} achieves
the shortest responses across all three backbone sizes, a combination no baseline achieves on any single model.
Token usage falls by \textbf{65.5\%} on the 1.5B model with an average accuracy gain of $+$4.53\%, and by \textbf{77.2\%} on Qwen3-8B with $+$5.33\%, ruling out scale as an explanation.
Figure~\ref{fig:buffer_dynamics} shows that accuracy and brevity improve in two distinct phases rather than competing with each other.
Batch accuracy climbs from 68.0\% to 76.5\% over 300 training steps while response length drops by 69.4\%.
The temporary accuracy dip around steps 100--150 marks the period of fastest length-target tightening, after which accuracy rebounds at a permanently lower token count.
This phased behaviour is built into the experience buffer: the length target $\ell^*_q$ is only tightened after the model has already produced a shorter correct answer, so the model is never pressured to be brief before it can be correct.

\textbf{(2) Larger models benefit more.}
IPT increases steadily with model size: from 7.23 to 23.29 on the 1.5B model,
from 11.15 to 32.84 on 7B, and from 8.64 to 51.04 on Qwen3-8B.
Stronger models find shorter correct solutions earlier in training, which tightens the experience buffer sooner and drives further compression.
On Qwen3-8B, MATH-500 responses shrink to just 755.7 tokens at 94.2\% accuracy, suggesting the model can solve most problems with very little text.

\textbf{(3) Fixed penalties cannot avoid the accuracy--brevity trade-off.}
Methods with aggressive length penalties, such as LC-R1, compress responses substantially but hurt accuracy; conservative methods such as Laser and JET preserve accuracy but stall near ${\sim}50\%$ or less compression.
The underlying reason is the same for both: without any memory of past training trajectories, the compression target stays fixed even as the model improves.
\ourMethod{} addresses this by recording the shortest correct solution found so far for each problem and using it as an adaptive target, reaching 65.5\% and 77.2\% token reduction on 1.5B and Qwen3-8B with no accuracy loss.
Among 1.5B baselines, AdaptThink is the closest competitor at $-$57.3\% tokens, yet its IPT of 18.04 is still 29\% below \ourMethod{}, and its Minerva accuracy of 22.1 trails \ourMethod{}'s 29.0, showing that even adaptive thresholds fall short without trajectory memory.

\begin{table}[t]
\centering
\caption{Pass@1 accuracy, average response length (Token), and Intelligence per Token (IPT) across all benchmarks. Vanilla denotes the unmodified base model. $\Delta$Acc and $\Delta$Token are average relative changes over Vanilla. \textbf{Bold}: best; \underline{underline}: second best.}
\label{tab:main_results}
\resizebox{\textwidth}{!}{%
\begin{tabular}{lccccccccccccc}
\toprule
\multirow{2}{*}{Method} & \multicolumn{2}{c}{AMC23} & \multicolumn{2}{c}{AIME24} & \multicolumn{2}{c}{MATH-500} & \multicolumn{2}{c}{Minerva} & \multicolumn{2}{c}{Olympiad} & \multicolumn{3}{c}{Avg} \\
\cmidrule(lr){2-3}\cmidrule(lr){4-5}\cmidrule(lr){6-7}\cmidrule(lr){8-9}\cmidrule(lr){10-11}\cmidrule(lr){12-14}
 & Acc & Token & Acc & Token & Acc & Token & Acc & Token & Acc & Token & $\Delta$Acc (\%) & $\Delta$Token (\%) & IPT \\
\midrule
\rowcolor[HTML]{DCE4EF}\multicolumn{14}{l}{\textit{DeepSeek-R1-Distill-Qwen-1.5B}} \\
Vanilla     & 62.0 & 8273.9  & 27.9 & 12019.2 & 81.0 & 4635.5 & 26.8 & 6347.1 & 41.5 & 8915.1  & --    & --      & 7.23  \\
LC-R1       & 61.3 & 4164.1  & 25.2 & 7359.4  & 78.6 & 2257.1 & 26.1 & 2383.3 & 39.6 & 4651.1  & -4.19 & -50.01  & 14.49 \\
Laser       & 60.2 & 4743.1  & 29.6 & 8044.2  & \textbf{84.6} & 2602.7 & \underline{28.7} & 3185.5 & \textbf{45.3} & 5121.3  & \textbf{+4.78} & -42.39  & 13.35 \\
AutoThink   & 60.0 & 4620.8  & 29.4 & 8108.2  & 79.8 & 2358.3 & 24.3 & 3201.2 & 42.4 & 5158.8  & -1.30 & -43.50  & 13.25 \\
AdaptThink  & 62.8 & \underline{3495.4}  & \textbf{30.4} & \underline{6704.7}  & \underline{83.8} & \underline{1851.1} & 22.1 & \underline{1968.5} & 44.6 & \underline{3978.2}  & +0.73 & \underline{-57.28}  & \underline{18.04} \\
JET         & \textbf{67.2} & 5671.4  & \underline{30.0} & 9730.5  & 82.8 & 3179.6 & 25.4 & 3988.1 & \underline{45.2} & 6490.0  & +4.37 & -29.25  & 10.86 \\
\rowcolor[HTML]{E3F1E8}\ourMethod  & \underline{65.8} & \textbf{2921.2}  & 28.8 & \textbf{5350.4}  & 82.2 & \textbf{1417.0} & \textbf{29.0} & \textbf{1761.0} & 43.0 & \textbf{3057.1}  & \underline{+4.53} & \textbf{-65.51}  & \textbf{23.29} \\
\midrule
\rowcolor[HTML]{DCE4EF}\multicolumn{14}{l}{\textit{DeepSeek-R1-Distill-Qwen-7B}} \\
Vanilla     & 78.9 & 6907.2  & 54.0 & 10063.7 & 89.8 & 3723.8 & 37.5 & 5017.3 & 54.4 & 7364.1  & --     & --      & 11.15 \\
LC-R1       & 76.4 & 3678.1  & 47.7 & 7002.4  & 89.0 & \underline{1529.3} & 32.4 & \underline{1734.4} & 54.2 & 4206.2  & -5.94  & -48.88  & \underline{23.47} \\
Laser       & \textbf{83.1} & \underline{3388.5}  & \underline{56.0} & \underline{6063.4}  & \underline{91.6} & 1868.2 & 37.1 & 2050.9 & \textbf{56.1} & \underline{3594.3}  & \textbf{+2.62}  & \underline{-50.17}  & 23.30 \\
AutoThink   & 81.6 & 4579.3  & 52.8 & 8247.3  & 91.3 & 2109.8 & 37.8 & 2617.4 & \underline{55.7} & 4968.7  & +1.21  & -35.09  & 18.63 \\
AdaptThink  & 79.1 & 4754.1  & 54.8 & 8693.6  & 91.0 & 1752.2 & 33.8 & 2645.4 & \underline{55.7} & 5603.0  & -0.88  & -33.78  & 19.52 \\
JET         & \underline{82.2} & 5573.4  & 52.3 & 9310.9  & 90.8 & 2971.1 & \textbf{40.4} & 3333.9 & 54.1 & 6230.3  & \underline{+1.87}  & -19.19  & 14.35 \\
\rowcolor[HTML]{E3F1E8}\ourMethod  & 80.4 & \textbf{2775.1}  & \textbf{54.4} & \textbf{5196.2}  & \textbf{92.4} & \textbf{1253.0} & \underline{38.6} & \textbf{1182.7}  & 54.7 & \textbf{2977.2}  & +1.80  & \textbf{-62.11}  & \textbf{32.84} \\
\midrule
\rowcolor[HTML]{DCE4EF}\multicolumn{14}{l}{\textit{Qwen3-8B}} \\
Vanilla     & 77.4 & 9139.1  & 61.7 & 11903.7 & 93.0 & 5315.2 & 44.1 & 7210.0 & 57.3 & 9649.2  & --     & --      & 8.64  \\

\rowcolor[HTML]{E3F1E8}\ourMethod  & \textbf{82.4} & \textbf{2232.0}  & \textbf{65.6} & \textbf{4581.3}  & \textbf{94.2} & \textbf{755.7}  & \textbf{45.5} & \textbf{850.4}  & \textbf{62.7} & \textbf{2430.0}  & \textbf{+5.33}  & \textbf{-77.18}  & \textbf{51.04} \\

\bottomrule
\end{tabular}%
}
\end{table}

\subsection{Ablation Studies}
\label{sec:ablation}

\begin{figure}[!t]
    \centering
    \includegraphics[width=0.98\linewidth]{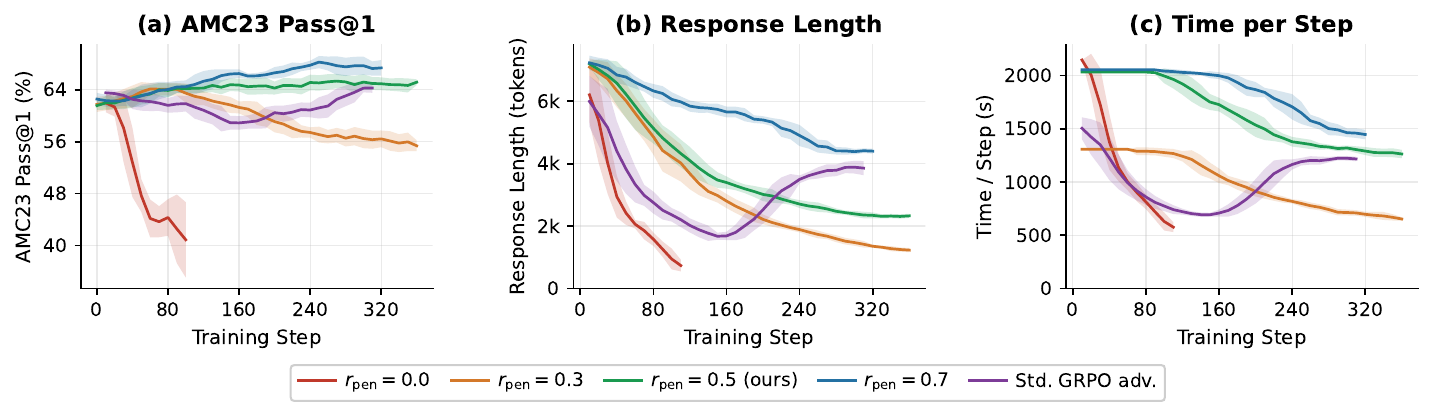}
    \caption{
        Training dynamics under different $r_{\text{pen}}$ settings and advantage functions.
        (a)~AMC23 Pass@1 over training steps.
        (b)~Average response length (tokens) over training steps.
        (c)~Wall-clock time per training step.
    }
    \label{fig:ablation_curves}
    \vspace{-10pt}
\end{figure}

We ablate the penalty coefficient $r_{\text{pen}}$ and the advantage normalization on the 1.5B model (Table~\ref{tab:ablation}, Figure~\ref{fig:ablation_curves}).


\paragraph{Discounted reward $r_{\text{pen}}$.}
With the harshest length penalty ($r_{\text{pen}}=0.0$), verbose but correct answers receive zero reward---identical to wrong answers---so the model learns to output very short but mostly wrong answers: AIME24 accuracy collapses to 7.92\% and MATH-500 to 43.6\%, far below the unmodified baseline.
Relaxing the penalty to 0.3 partially restores accuracy but is still too aggressive to maintain the accuracy-first training progression needed for stable learning.
The default value of 0.5 strikes the best balance, yielding MATH-500 accuracy of 82.2\% at 1,417 tokens and AIME24 accuracy of 28.8\% at 5,350 tokens with an average IPT of 23.3, as reflected in the stable training curves in Figure~\ref{fig:ablation_curves}(a) and (b).
A weaker penalty of $r_{\text{pen}}=0.7$ raises accuracy on most benchmarks by 2--7 percentage points but reduces Minerva accuracy by 2.9 points, increases average response length by over 50\%, and cuts IPT by roughly 37\% (Figure~\ref{fig:ablation_curves}(c)), making it impractical under standard compute budgets.

\begin{table}[t]
\small
\centering
\caption{Ablation studies on the 1.5B model. Green row marks the default \ourMethod{} setting.}
\label{tab:ablation}
\resizebox{0.98\textwidth}{!}{%
\begin{tabular}{llccccccccccl}
\toprule
\multicolumn{2}{l}{\multirow{2}{*}{Setting}} & \multicolumn{2}{c}{AMC23} & \multicolumn{2}{c}{AIME24} & \multicolumn{2}{c}{MATH-500} & \multicolumn{2}{c}{Minerva} & \multicolumn{2}{c}{Olympiad} & \multirow{2}{*}{Avg IPT} \\
\cmidrule(lr){3-4}\cmidrule(lr){5-6}\cmidrule(lr){7-8}\cmidrule(lr){9-10}\cmidrule(lr){11-12}
 & & Acc & Token & Acc & Token & Acc & Token & Acc & Token & Acc & Token & \\
\midrule
\rowcolor[HTML]{EBEBEB}\multicolumn{13}{l}{\textit{Ablation on discounted reward $r_{\text{pen}}$}} \\
\multicolumn{2}{l}{$r_{\text{pen}}=0.0$} & 34.5 & 735.7  & 7.92 & 1510.4 & 43.6 & 313.8  & 18.4 & 224.3  & 25.5 & 751.6  & 61.4 \\
\multicolumn{2}{l}{$r_{\text{pen}}=0.3$} & 55.7 & 1510.6 & 16.9 & 2780.7 & 70.4 & 650.7  & 24.6 & 601.6  & 35.1 & 1484.7 & 43.1 \\
\rowcolor[HTML]{E3F1E8}\multicolumn{2}{l}{$r_{\text{pen}}=0.5$ \textbf{(ExpThink)}} & \textbf{65.8} & \textbf{2921.2} & \textbf{28.8} & \textbf{5350.4} & \textbf{82.2} & \textbf{1417.0} & \textbf{29.0} & \textbf{1761.0} & \textbf{43.0} & \textbf{3057.1} & \textbf{23.3} \\
\multicolumn{2}{l}{$r_{\text{pen}}=0.7$} & 67.9 & 4517.1 & 35.8 & 7556.9 & 84.8 & 2343.2 & 26.1 & 3103.9 & 46.5 & 4969.8 & 14.7 \\
\rowcolor[HTML]{EBEBEB}\multicolumn{13}{l}{\textit{Ablation on advantage function ($r_{\text{pen}}=0.5$)}} \\
\multicolumn{2}{l}{Std.\ GRPO adv.} & 63.5 & 5334.2 & 35.4 & 8696.3 & 77.4 & 1870.4 & 22.9 & 3226.1 & 43.5 & 5326.6 & 14.5 \\
\bottomrule
\end{tabular}%
}
\end{table}

\paragraph{Count-based advantage normalization.}
When the correct-count $|\mathcal{C}_q|$ is replaced by the standard deviation $\sigma$ as the advantage denominator (Std.\ GRPO adv.), accuracy stays similar but the model generates 1.6 to 1.8 times more tokens per response, dropping average IPT from 23.3 to 14.5; response length never reaches the compact levels seen in \ourMethod{} (Figure~\ref{fig:ablation_curves}(b)).
The reason is that $\sigma$ does not distinguish between easy and hard problems: it assigns similar gradient magnitudes to both, so the model receives no signal to be brief on easy questions and no extra push to be correct on hard ones, breaking the accuracy-first pattern that \ourMethod{} relies on.

\subsection{Dissecting Compressed Reasoning Traces}
\label{sec:trace}

\paragraph{Overthinking keyword suppression.}

\begin{figure}[!ht]
    \centering
    \begin{minipage}[t]{0.495\linewidth}
        \centering
        \includegraphics[width=\linewidth]{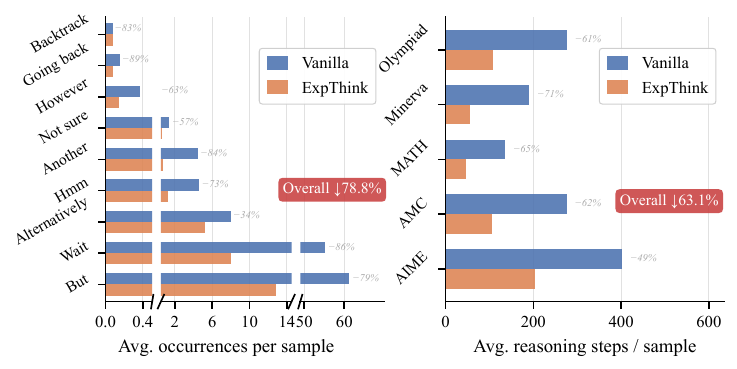}
        \centerline{\small (a)}
    \end{minipage}
    \hfill
    \begin{minipage}[t]{0.495\linewidth}
        \centering
        \includegraphics[width=\linewidth]{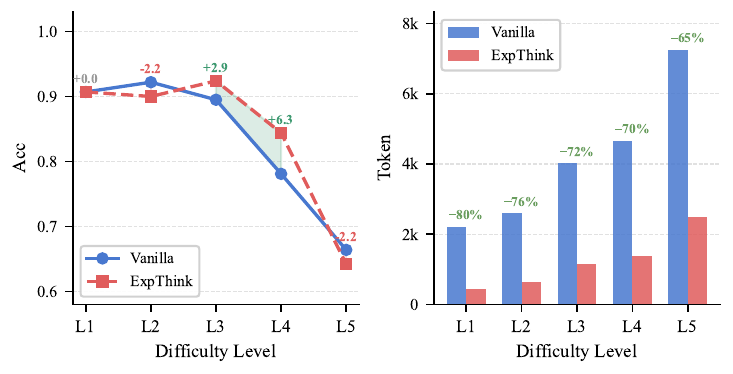}
        \centerline{\small (b)}
    \end{minipage}
    \caption{
        Analysis of \textbf{ExpThink}'s behaviour.
        (a)~Overthinking suppression across keywords and datasets;
        (b)~Difficulty-adaptive behaviour on MATH-500.
    }
    \label{fig:analysis}
\end{figure}

We segment reasoning traces at double-newline boundaries and count
overthinking indicators following~\citet{lu2025retro}:
\{\textit{But, Wait, Alternatively, However, Hmm, Not sure, Going back,
Backtrack, Another}\}, which are lexical markers of hesitation,
self-correction, or exploratory branching.
As shown in Figure~\ref{fig:analysis}(a-left), \ourMethod reduces
the total frequency of these keywords by \textbf{78.8\%}, with \textit{But} and
\textit{Wait}, the two most common ones, falling by 79\% and 86\% respectively, while accuracy improves from 50.3\% to 52.3\%.
The number of reasoning steps also drops by \textbf{63.1\%},
from 218.3 to 80.6 steps on average (Figure~\ref{fig:analysis}(a-right)).
The reduction is steeper on Minerva than on AIME, where problems genuinely require longer derivations.
This pattern shows that \ourMethod primarily cuts unnecessary hedging and backtracking, rather than shortening all responses uniformly.

\paragraph{Difficulty-adaptive compression.}

Figure~\ref{fig:analysis}(b) reports accuracy and response length
across the five difficulty tiers of MATH-500.
As expected, easier problems are compressed more aggressively: Level-1 responses shrink by 79.8\% while Level-5 shrinks by 65.4\%.
This happens because easy problems are solved correctly by more rollouts, giving them a larger $|\mathcal{C}_q|$ that weakens the advantage signal and pushes the model toward brevity.
For harder problems where fewer rollouts succeed, the stronger advantage signal prioritises getting the answer right over being short, which is why accuracy gains are larger at higher difficulty: Level-4 improves by 6.3 percentage points and Level-3 by 2.9 percentage points.
Even so, compression remains substantial at the top tier: Level-5 saves approximately 4,740 tokens per response on average, compared with roughly 1,767 at Level-1.

\subsection{Trajectory Diversity and Inference-Time Scaling}

\label{sec:budget}

\begin{figure}[!ht]
    \centering
    \begin{minipage}[t]{0.495\linewidth}
        \centering
        \includegraphics[width=\linewidth]{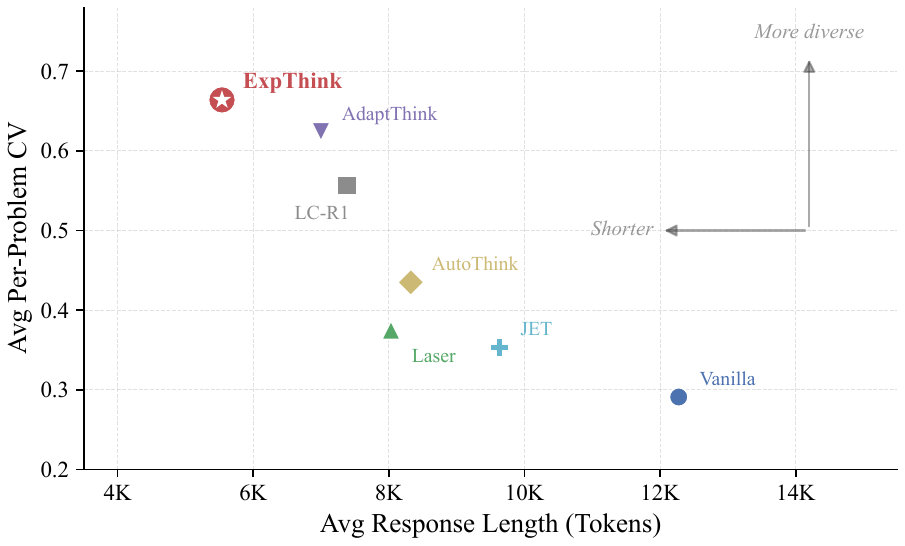}
        \centerline{\small (a) Trajectory diversity across methods}
    \end{minipage}
    \hfill
    \begin{minipage}[t]{0.495\linewidth}
        \centering
        \includegraphics[width=\linewidth]{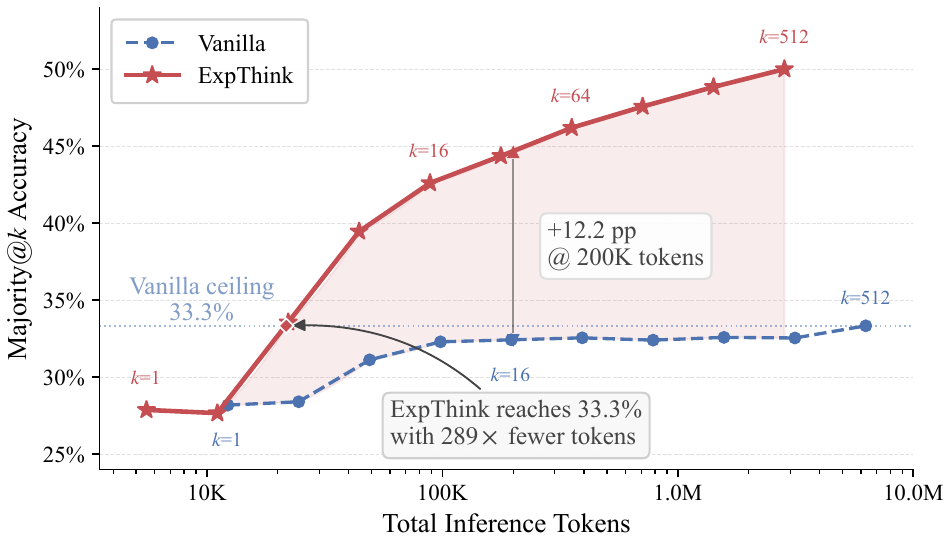}
        \centerline{\small (b) Majority@$k$ scaling under token budget}
    \end{minipage}
    \caption{
        Inference-time analysis on AIME24.
        (a)~Average response length vs.\ per-problem coefficient of variation (CV) of response length. 
        (b)~Majority@$k$ accuracy versus total inference tokens consumed.DeepSeek-R1-Distill-Qwen-1.5B, 512 samples per problem.
    }
    \label{fig:exp-efficiency}
\end{figure}

Shorter traces allow more parallel rollouts within a fixed compute budget.
We sample $k \in \{1, 2, 4, \ldots, 512\}$ independent responses per problem on AIME24 and report majority-voting accuracy against total tokens consumed.

\paragraph{Diversity of reasoning trajectories.}
To understand \emph{why} \ourMethod scales better with repeated sampling, we measure the per-problem coefficient of variation (CV) of response length across 512 rollouts, which captures how much a model's solution length varies from one attempt to the next.
As shown in Figure~\ref{fig:exp-efficiency}(a), \ourMethod has the highest CV at 0.664 with an average response length of only 5.5K tokens, while the Vanilla baseline produces consistently long outputs averaging 12.3K tokens and a much lower CV of 0.291.
In other words, \ourMethod sometimes finds a short solution and sometimes takes longer, exploring varied reasoning paths rather than always producing responses of similar length.

\paragraph{Majority-voting scaling.}
This diversity translates directly into better performance under a fixed token budget.
As shown in Figure~\ref{fig:exp-efficiency}(b), the Vanilla baseline stops improving at around 33.3\% accuracy from $k = 16$ onward, even when up to 6.3 million tokens are spent on additional samples.
\ourMethod keeps improving with more samples, reaching 50.0\% at $k = 512$.
Given the same budget of 200K tokens, \ourMethod outperforms Vanilla by 12.2 percentage points; put differently, it reaches Vanilla's peak accuracy with 289 times fewer tokens.
Since \ourMethod's shorter responses allow more independent attempts within any fixed budget, and each attempt explores a different reasoning path, additional samples continue to add value rather than repeating the same answer.

\subsection{Generalization to out-of-distribution benchmarks.}
\label{sec:generalization}

\begin{table}[t]
\centering
\caption{Out-of-domain generalization results. Pass@1 accuracy, average response length (Token), and Intelligence per Token (IPT) on non-math benchmarks. \(\Delta\)Acc and \(\Delta\)Token denote the average relative change with respect to the Vanilla baseline.}
\label{tab:ood_results}
\resizebox{\textwidth}{!}{%
\begin{tabular}{lccccccccccc}
\toprule
\multirow{2}{*}{Method} & \multicolumn{2}{c}{LiveCodeBench} & \multicolumn{2}{c}{GPQA-Diamond} & \multicolumn{2}{c}{MMLU} & \multicolumn{3}{c}{Avg} \\
\cmidrule(lr){2-3}\cmidrule(lr){4-5}\cmidrule(lr){6-7}\cmidrule(lr){8-10}
 & Acc & Token & Acc & Token & Acc & Token & $\Delta$Acc (\%) & $\Delta$Token (\%) & IPT \\
\midrule
\rowcolor[HTML]{DCE4EF}\multicolumn{10}{l}{\textit{DeepSeek-R1-Distill-Qwen-1.5B}} \\
Vanilla       & 12.80 & 6218   & 16.0          & 8514.3         & 38.8          & 1680.3         & --              & --              & 9.01           \\
LC-R1         & 9.48  & \textbf{3217}   & 23.0          & 5285.2         & 41.0          & \underline{927.0}  & +7.83           & \underline{-43.67}       & 17.18          \\
Laser         & \textbf{13.93} & 6389   & 21.8          & 5136.9         & 42.5          & 943.5          & +18.21          & -26.92          & 17.16          \\
AutoThink     & 12.61 & 5870   & \underline{25.0}         & 5324.0         & 40.5          & 1068.1         & +19.72          & -26.50          & 14.92          \\
AdaptThink    & 13.55 & 6355   & 24.4          & \underline{4013.8}        & \underline{44.7}         & 995.2          & \underline{+24.52}       & -30.48          & \underline{17.71}        \\
JET           & 11.56 & 6820   & 23.9          & 6566.8         & \textbf{46.7}         & 1027.1         & +20.02          & -17.35          & 16.93          \\
\rowcolor[HTML]{E3F1E8}\ourMethod    & \underline{13.65} & \underline{4626}   & \textbf{25.4}         & \textbf{3259.4}        & 44.2          & \textbf{622.1}         & \textbf{+26.44}          & \textbf{-50.10}          & \textbf{27.26}          \\
\midrule
\rowcolor[HTML]{DCE4EF}\multicolumn{10}{l}{\textit{DeepSeek-R1-Distill-Qwen-7B}} \\
Vanilla       & 19.15 & 6509   & 35.2          & 7288.7         & 63.9          & 1170.8         & --              & --              & 20.78          \\
LC-R1         & 16.11 & \underline{5123}   & 40.8          & 4295.6         & 63.4          & \underline{649.6}  & -0.25           & -35.62          & \underline{36.75}        \\
Laser         & \underline{21.04} & 5153   & \underline{44.1}         & \underline{3492.9}        & \textbf{65.1}         & 781.1          & \underline{+12.34}       & \underline{-50.17}       & 33.35          \\
AutoThink     & \textbf{21.99} & 5912   & 37.9          & 5431.5         & 57.4          & 861.0          & +4.11           & -20.37          & 25.79          \\
AdaptThink    & 20.28 & 5980   & 38.3          & 5225.0         & 63.2          & 902.6          & +4.54           & -19.78          & 26.91          \\
JET           & 20.19 & 6156   & 40.3          & 6491.1         & \underline{64.6}         & 1036.0         & +7.00           & -9.29           & 23.95          \\
\rowcolor[HTML]{E3F1E8}\ourMethod    & 20.73 & \textbf{4538}   & \textbf{47.2}         & \textbf{2870.9}        & \textbf{65.1}         & \textbf{448.5}         & \textbf{+14.74}          & \textbf{-50.86}          & \textbf{55.39}          \\
\bottomrule
\end{tabular}%
}
\end{table}

Table~\ref{tab:ood_results} complements the in-domain results in Table~\ref{tab:main_results} by testing whether the efficiency gains on math transfer to other task types without any further fine-tuning.
On the 1.5B model, \ourMethod achieves the largest accuracy improvement and the greatest token reduction among all methods, with an IPT of 27.26 compared with 9.01 for the Vanilla baseline.
The gains are most pronounced on GPQA-Diamond, where accuracy rises from 16.0 to 25.4, and on MMLU, where it rises from 38.8 to 44.2; these are exactly the types of knowledge-intensive tasks where models tend to over-deliberate, so the conciseness training is especially effective.
On the 7B model, \ourMethod reaches an IPT of 55.39, nearly 2.7 times the Vanilla baseline's 20.78, and achieves the best GPQA-Diamond accuracy of 47.2 among all methods.
Overall, these results show that the compression behaviour learned on math problems carries over to code and science tasks, reflecting a general reduction in unnecessary reasoning steps rather than a math-specific shortcut.

\section{Conclusion}
\label{sec:conclusion}

We introduced \ourMethod, a reinforcement learning framework that breaks the epoch-independence assumption of prior CoT compression methods. By maintaining a per-problem experience buffer of the shortest correct solution observed across training, \ourMethod defines an adaptive reward threshold that tightens automatically as the policy improves, eliminating manual schedule design. Paired with count-based advantage normalization, this self-evolving curriculum allocates optimization pressure toward correctness on hard problems and toward brevity once correctness is reliable. Across three backbones, \ourMethod consistently improves the accuracy--efficiency tradeoff, reducing token usage by up to 77\% while simultaneously gaining accuracy, and the gains generalize to out-of-distribution benchmarks including GPQA-Diamond and LiveCodeBench. These results establish cross-epoch trajectory reuse as a lightweight and effective principle for training reasoning models with higher intelligence per token.

\paragraph{Limitations.}
The training data and primary evaluation are limited to mathematical reasoning with verifiable answers, and transfer to interactive tool use or multimodal settings remains to be confirmed. The reward also depends on reliable answer verification, which may not hold for tasks with ambiguous or partially correct outputs. A controlled comparison against DAPO under matched compute, which would further disentangle the contributions of the experience buffer from those of DAPO-style training, is also left to future work. Finally, the tolerance margin $\alpha$ and discounted reward $r_{\text{pen}}$ that guard against over-compression are empirical choices; although $r_{\text{pen}}$ is ablated in Section~\ref{sec:ablation}, a systematic study of $\alpha$ sensitivity across diverse domains is left to future work.

\paragraph{Future work.}
We plan to extend \ourMethod to multimodal reasoning, where concise traces must preserve visual grounding. A further direction is to store full trajectories in the experience buffer and perform off-policy replay, enabling the curriculum to leverage not only length statistics but also the content of past successful solutions for richer reward shaping and diversity-aware exploration.

\clearpage

\bibliographystyle{plainnat}
\bibliography{references}

\clearpage
\appendix

\section{Related Work}
\label{sec:related}

\subsection{Experience-Guided Reinforcement Learning with Verifiable Rewards}

A growing body of work leverages historical rollout experience to improve GRPO training. On the experience-reuse side, LANPO~\cite{li2025lanpo} accumulates past rollouts into a dynamic pool for cross-sample abstraction, BAPO~\cite{wan2026buffer} replays difficulty-aware batches of previously failed problems, RPO~\cite{yi2026rpo} caches and reuses trajectory prefixes to cut rollout cost by up to 95\%, and SOUP~\cite{yang2026soup} splices off-policy prefixes with on-policy continuations via importance weighting. On the reward-shaping side, WS-GRPO\cite{mundada2026ws} learns prefix-level continue/stop signals from outcome-only data, F-GRPO~\cite{plyusov2026f} applies focal-loss-inspired re-weighting to shift gradient mass toward rare-correct trajectories, XRPO~\cite{bamba2025xrpo} sharpens advantages via novelty-aware rollout allocation, HiLL~\cite{xia2026learning} co-trains a hinter policy to recover learning signals on hard problems, and STEP~\cite{chen2025step} drives adaptive resampling from historical success rates.

\subsection{Reinforcement Learning Based Efficient Reasoning}

Reinforcement learning offers a more direct route by jointly rewarding correctness and brevity, allowing the model to discover shorter valid paths without annotated demonstrations.
The dominant paradigm augments the reward with a length penalty, as in L1~\cite{aggarwal2025l1}, ConciseRL~\cite{dumitru2025conciserl}, ThinkPrune~\cite{hou2025thinkprune}, and LC-R1~\cite{cheng2025optimizing}.
DLER~\cite{liu2025dler} identifies that naive length penalties can suppress high-entropy exploration tokens critical for reasoning, proposing a recipe that recovers accuracy while reducing response length by over 70\%. LEASH~\cite{li2025leash} introduces an adaptive Lagrangian penalty that dynamically adjusts the constraint coefficient to accelerate convergence toward a target length.
GR3~\cite{li2026tackling} replaces additive length penalties with multiplicative group-relative reward rescaling, eliminating compensatory optimization shortcuts that arise in standard formulations.
Dr.\ GRPO~\cite{liu2025drgrpo} corrects an optimization bias in GRPO that artificially inflates response length for incorrect outputs. Adaptive frameworks such as DAST and AutoThink~\cite{shen2025dast,tu2025learning,zhang2025adaptthink} allocate inference budgets as a function of estimated problem difficulty, and Laser~\cite{liu2025learn} combines difficulty-aware penalties with explicit exploration of incorrect trajectories.
ProRL v2~\cite{nvidia2025prorl} demonstrates that scheduled cosine length penalties within prolonged RL training enable sustained improvement over thousands of steps without the model collapsing to short contexts.
Beyond scalar penalties, finer-grained formulations have emerged: SWAP~\cite{li2026stepwise} applies step-level penalties weighted by each step's contribution to the correct answer; FGO~\cite{han2026long} subdivides GRPO response groups by length and entropy to prevent representation collapse during compression; SCMA~\cite{chen2026self} uses a dual-agent RL loop that segments reasoning into chunks and scores each for redundancy; CEEH~\cite{luo2026compress} selectively regularizes entropy based on estimated problem difficulty; and \citet{li2025reinforcement} tie length penalties to whether the model has already demonstrated ability to solve a given problem with a shorter response, enabling cross-domain generalization.
Despite their diversity, all these RL methods treat each training epoch independently: the reward at epoch $e$ depends only on trajectories sampled during epoch $e$, leaving cross-epoch compression progress unused.

\section{Training Hyperparameters}
\label{app:training_params}

We train all models on 8$\times$ A800 80GB GPUs with a global batch size of 512 and a maximum sequence length of 16{,}384 tokens. To address memory constraints under long-context training, we employ Fully Sharded Data Parallel (FSDP) with CPU offloading.

During optimization, we adopt the GRPO objective with $\beta = 0.001$ and asymmetric clipping parameters $\varepsilon_{\text{low}} = 0.2$ and $\varepsilon_{\text{high}} = 0.28$. The learning rate is fixed at $1 \times 10^{-6}$. We use AdamW as the optimizer with default parameter settings. For response generation during training, we use temperature $= 1.0$ and top-$p = 1.0$ to encourage exploration; during validation, we use temperature $= 0.6$ and top-$p = 0.95$. Training runs for 300 update steps. At each step, we sample $G = 16$ responses per prompt to construct response groups for GRPO-based advantage computation. Validation is conducted periodically during training, and checkpoints are saved every 10 steps for model selection and reproducibility.

Notably, we do not incorporate KL divergence regularization into the reward function; the reward signal is computed independently of any KL penalty term.

\section{Datasets}
\label{app:datasets}

\paragraph{DeepScaleR~\citep{luo2025deepscaler}.}
DeepScaleR is a curated mathematical reasoning dataset comprising approximately 40K problems, constructed by scaling up the difficulty distribution of existing open-source mathematical corpora. It is built upon a systematic difficulty-aware selection pipeline: problems are first scored by estimated difficulty, and then progressively sampled to emphasize challenging instances that demand multi-step derivations, algebraic manipulation, and advanced problem-solving strategies. DeepScaleR serves as the training corpus for \ourMethod. Its main strengths lie in its carefully calibrated difficulty gradient, broad coverage of mathematical topics spanning algebra, geometry, number theory, and combinatorics.

\paragraph{AIME24~\citep{codeforcesamerican}.}
AIME24 consists of the official problems from the 2024 American Invitational Mathematics Examination (AIME) I and II and has become a widely used benchmark for evaluating high-level mathematical reasoning. Its problems typically require sophisticated combinatorial analysis, geometric constructions, or number-theoretic reasoning, and each admits a unique three-digit integer answer. In public repositories, AIME24 usually appears as a 30-problem set designed to assess models under realistic contest conditions.

\paragraph{AMC23~\citep{amc23}.}
AMC23 is a benchmark derived from the 2023 American Mathematics Competition, primarily AMC 10 and AMC 12. The problems are reformatted for model evaluation and usually take the form of multiple-choice or integer-answer questions, covering core high-school contest topics such as algebra, geometry, probability, and number theory. The overall difficulty exceeds routine classroom exercises but remains below that of AIME and Olympiad-level problems. Compared with AIME24, AMC23 typically contains a larger number of instances and is therefore commonly used as a medium-difficulty benchmark.

\paragraph{MATH-500~\citep{hendrycks2021measuring}.}
MATH-500 is a curated subset of 500 competition-level problems drawn from the MATH dataset. The original MATH dataset contains approximately 12{,}500 U.S.\ high-school competition problems, each accompanied by a full step-by-step solution, and spanning topics such as algebra, geometry, counting and probability, number theory, and precalculus. Difficulty levels follow the Art of Problem Solving (AoPS) scale from 1 to 5. Evaluation is conducted using an exact-match criterion on the final answer.

\paragraph{Minerva Math~\citep{lewkowycz2022solving}.}
Minerva Math is a collection of mathematical and quantitative reasoning problems assessing formal mathematical and scientific reasoning capabilities. In publicly released evaluations, roughly 272 problems are included, primarily sourced from MIT OpenCourseWare and covering advanced undergraduate- and graduate-level STEM topics such as solid-state chemistry, astronomy, differential equations, and special relativity. Compared with pure mathematics competition datasets, Minerva Math emphasizes cross-disciplinary quantitative problems.

\paragraph{OlympiadBench~\citep{he2024olympiadbench}.}
OlympiadBench is a bilingual multimodal Olympiad-level science benchmark designed to evaluate the reasoning and problem-solving capabilities of LLMs using extremely challenging mathematics and physics competition problems. The dataset aggregates 8{,}476 questions drawn from international and national mathematics Olympiads, physics contests, and the Chinese college entrance examination, spanning both text-only and image-based multimodal formats. Empirical results indicate that even state-of-the-art systems achieve relatively low performance, highlighting the benchmark's exceptional difficulty.

\paragraph{MMLU~\citep{hendrycks2020measuring}.}
MMLU (Massive Multitask Language Understanding) is a comprehensive benchmark spanning 57 diverse academic subjects, ranging from STEM disciplines to humanities and social sciences. Each problem is presented in a four-option multiple-choice format, and the questions are drawn from real-world examination materials at difficulty levels from elementary to advanced professional. MMLU is used in our work to assess out-of-distribution generalization of the efficiency learned by \ourMethod.

\paragraph{LiveCodeBench~\citep{jain2024livecodebench}.}
LiveCodeBench is a holistic and contamination-free benchmark designed to evaluate
large language models on a broad spectrum of code-related capabilities. It
continuously collects problems from competitive programming platforms---LeetCode,
AtCoder, and CodeForces---using only contests published \emph{after} the training
cutoffs of most evaluated models, thereby eliminating data-leakage artifacts that
plague static coding benchmarks such as HumanEval and MBPP. Each problem falls
into one of four task categories: code generation, self-repair, code execution, and
test-output prediction. LiveCodeBench is used in our work to evaluate whether the
reasoning efficiency acquired through \ourMethod generalizes to complex,
out-of-distribution programming tasks.

\paragraph{GPQA~\citep{rein2024gpqa}.}
GPQA (Graduate-Level Google-Proof Q\&A) is a rigorous multiple-choice benchmark
comprising 448 questions across biology, physics, and chemistry, authored by
domain specialists with doctoral-level expertise. The questions are deliberately
crafted to resist web-search assistance: skilled non-experts who spent over 30
minutes searching the internet achieved only 34\% accuracy, while in-domain PhD
experts reached approximately 65\%, underscoring the benchmark's exceptional
difficulty. GPQA is used in our work to probe whether \ourMethod improves deep
scientific reasoning in settings that demand genuine domain knowledge rather than
surface-level pattern matching.

\section{Detailed Evaluation Protocol}
\label{app:eval_protocol}

\paragraph{Accuracy (Acc).}
For each benchmark dataset, we measure accuracy as the proportion of correctly answered problems:
\begin{equation}
    \text{Acc} = \frac{1}{N} \sum_{i=1}^{N} \mathbf{1}(\hat{y}_i = y_i),
\end{equation}
where $N$ is the total number of problems, $\hat{y}_i$ is the model's predicted answer, $y_i$ is the ground-truth answer, and $\mathbf{1}(\cdot)$ is the indicator function.

\paragraph{Average Token Length (Token).}
We report the average number of tokens generated by the model across all problems in a dataset:
\begin{equation}
    \text{Token} = \frac{1}{N} \sum_{i=1}^{N} l_i,
\end{equation}
where $l_i$ is the number of tokens in the model's response for the $i$-th problem.

\paragraph{Intelligence per Token (IPT).}
To jointly evaluate accuracy and computational cost, we define Intelligence per Token (IPT) as:
\begin{equation}
    \text{IPT} = \frac{\text{Acc}}{\text{Token} / 1000},
\end{equation}
where Acc denotes the Pass@1 accuracy (\%) and Token denotes the average output token length. Intuitively, IPT quantifies the accuracy gained per 1K tokens of generation, providing a unified measure of intelligence density. A higher IPT indicates that the model achieves stronger reasoning performance with fewer tokens, reflecting a more favorable accuracy--efficiency trade-off. In the main results (Table~\ref{tab:main_results}), we report the IPT value averaged across all benchmark datasets for each method.

\paragraph{Average Change Rate ($\Delta$).}
To quantify the variation in performance introduced by a method relative to the original base model, we define the change rate of a metric $M$ on a single dataset as:
\begin{equation}
    \text{CR}(M) = \frac{M_{\text{Method}}}{M_{\text{Vanilla}}} - 1,
\end{equation}
where $M \in \{\text{Acc}, \text{Token}\}$. The Average Change Rate aggregates the change rates across all $K = 5$ evaluation datasets:
\begin{equation}
    \Delta(M) = \frac{1}{K} \sum_{j=1}^{K} \text{CR}_j(M).
\end{equation}

\paragraph{Inference setup.}
During inference, the maximum context size is limited to 16{,}384 tokens, and the
sampling temperature is fixed at 0.6. For the AIME dataset, due to its relatively
small sample size, we repeat the evaluation 16 times and report the averaged
results to reduce variance and improve statistical reliability. Similarly, for the
AMC dataset, we repeat the evaluation 8 times and report the averaged results.

\section{Detailed Descriptions of Baselines}
\label{app:baselines}

All baseline methods are grounded in reinforcement learning (RL) and address the problem of efficient reasoning from distinct perspectives. 


\paragraph{LC-R1~\citep{cheng2025optimizing}.}
LC-R1 (Length-Compressed R1) addresses reasoning efficiency by augmenting the GRPO objective with two complementary reward signals. The first is an explicit \emph{length reward} that penalizes the total token count of generated reasoning sequences, applying a global conciseness pressure. The second is a \emph{compression reward} that specifically targets the elimination of redundant self-verification patterns—i.e., segments where the model unnecessarily rechecks already-established conclusions—without broadly suppressing productive reasoning steps. This targeted decomposition allows LC-R1 to achieve approximately 50\% reduction in sequence length with only around 2\% accuracy degradation. We use the publicly released checkpoints LCR1-1.5B and LCR1-7B.

\paragraph{Laser~\citep{liu2025learn}.}
Laser (Length-bAsed StEp rEward shaping) provides a unified framework for efficient reasoning through step-level length-based reward shaping, using a step function controlled by a target length as the primary reward signal. Its extended variant, \textbf{LASER-D}, incorporates two additional mechanisms: \emph{dynamic adaptation}, where the reward specification evolves continuously as model behavior changes throughout training; and \emph{difficulty-awareness}, which imposes heavier length penalties on easier queries to encourage fast responses for simple problems while preserving extended chain-of-thought for genuinely hard ones. This explicit difficulty-conditioned reward design allows the model to learn a problem-adaptive reasoning budget. We use the checkpoints Laser-DE-L4096-1.5B and Laser-DE-L4096-7B, both trained on the same dataset as ours.

\paragraph{AutoThink~\citep{tu2025learning}.}
AutoThink is motivated by the observation that inserting a simple ellipsis token into prompts can stochastically trigger either thinking or no-thinking mode in R1-style models, revealing a latent controllability in reasoning behavior. Building on this finding, AutoThink proposes a three-stage reinforcement learning curriculum with progressive reward shaping. Stage~1 employs batch-level reward balancing to maintain the coexistence of both thinking and no-thinking modes during early training; Stage~2 reinforces task accuracy within each mode independently; Stage~3 introduces a length-aware reward to suppress unnecessary computation and encourage direct responses for simpler queries. Trained on the same dataset as ours, we use the checkpoints Distill-R1-1.5B-AutoThink-Stage3 and Distill-R1-7B-AutoThink-Stage3.

\paragraph{AdaptThink~\citep{zhang2025adaptthink}.}
AdaptThink is grounded in the empirical finding that forcing reasoning models to always engage in extended chain-of-thought is suboptimal: for simpler problems, skipping CoT entirely (\emph{NoThinking} mode) can improve both speed and accuracy. AdaptThink leverages RL with two complementary components to exploit this observation: a \emph{constrained optimization} objective that nudges the model toward NoThinking responses while preserving overall accuracy, and an \emph{importance-sampling} scheme that balances the ratio of Thinking versus NoThinking training samples to enable stable exploration of both modes. Together, these enable the model to adaptively select whether to invoke full chain-of-thought reasoning or to respond directly, based on perceived problem difficulty. Trained on the same dataset as ours, we use the checkpoints AdaptThink-1.5B-delta0.05 and AdaptThink-7B-delta0.05.

\paragraph{JET~\citep{han2025your}.}
JET (Just-Enough Thinking) is motivated by Evidence Accumulation Models from cognitive science, which suggest that sufficient evidence for a decision is often accumulated early in a reasoning sequence, rendering subsequent steps redundant. JET operationalizes this insight through two mechanisms: \emph{trajectory truncation during rollout}, which exposes the model to shorter, distributionally consistent reasoning paths during RL training to overcome the distributional barrier that prevents naive length rewards from generating compact rollouts; and a \emph{quality-controlled length reward}, which incentivizes conciseness while explicitly preserving answer correctness. Rather than penalizing long outputs after generation, JET shapes the model's internal stopping criterion so that it proactively terminates reasoning at the appropriate point. We use the checkpoints JET-1.5B and JET-7B. JET has been accepted at ICLR 2026.

\section{Supplementary Experiments}
\label{app:supplementary}

\subsection{Difficulty-Aware Accuracy Analysis}
\label{app:difficulty_acc}

\begin{table}[h]
\centering
\caption{Accuracy comparison between Vanilla and \ourMethod across difficulty levels on MATH-500 (1.5B model). Difficulty quintiles are defined by Vanilla response length, where Q1 (shortest) represents easy problems and Q5 (longest) represents the hardest problems that push the Vanilla model toward its context limit.}
\label{tab:difficulty_acc}
\begin{tabular}{lcccccc}
\toprule
Method & Q1 (Easy) & Q2 & Q3 & Q4 & Q5 (Hard) & Overall \\
\midrule
Vanilla    & 0.940 & 0.930 & 0.960 & 0.870 & 0.410 & 0.822 \\
\ourMethod & \textbf{0.940} & \textbf{0.960} & 0.930 & \textbf{0.800} & \textbf{0.520} & \textbf{0.830} \\
\midrule
Token $\Delta$ & $-$60.0\% & $-$66.9\% & $-$61.9\% & $-$67.3\% & $-$73.3\% & $-$66.6\% \\
\bottomrule
\end{tabular}
\end{table}

The results reveal a key pattern in \ourMethod's difficulty-adaptive behavior. On easy problems (Q1--Q3), both models achieve similar accuracy, but \ourMethod uses 60--67\% fewer tokens, demonstrating efficient compression without quality loss. On the hardest problems (Q5), \ourMethod achieves substantially higher accuracy (52.0\% vs.\ 41.0\%, a relative gain of $+$26.8\%) while simultaneously using 73.3\% fewer tokens. This is precisely the behavior intended by the difficulty-adaptive gradient reweighting: easy problems receive suppressed gradients (compression-first), while hard problems receive amplified gradients (accuracy-first). The token reduction is actually largest on the hardest quintile, because many Vanilla responses on hard problems hit the 16{,}384-token context limit and still fail—\ourMethod finds shorter, more direct solution paths that avoid the error-inducing overthinking characteristic of the Vanilla model.

\subsection{Ablation Study Training Curves}
\label{app:ablation_curves}

\begin{figure}[h]
  \centering
  \includegraphics[width=\linewidth]{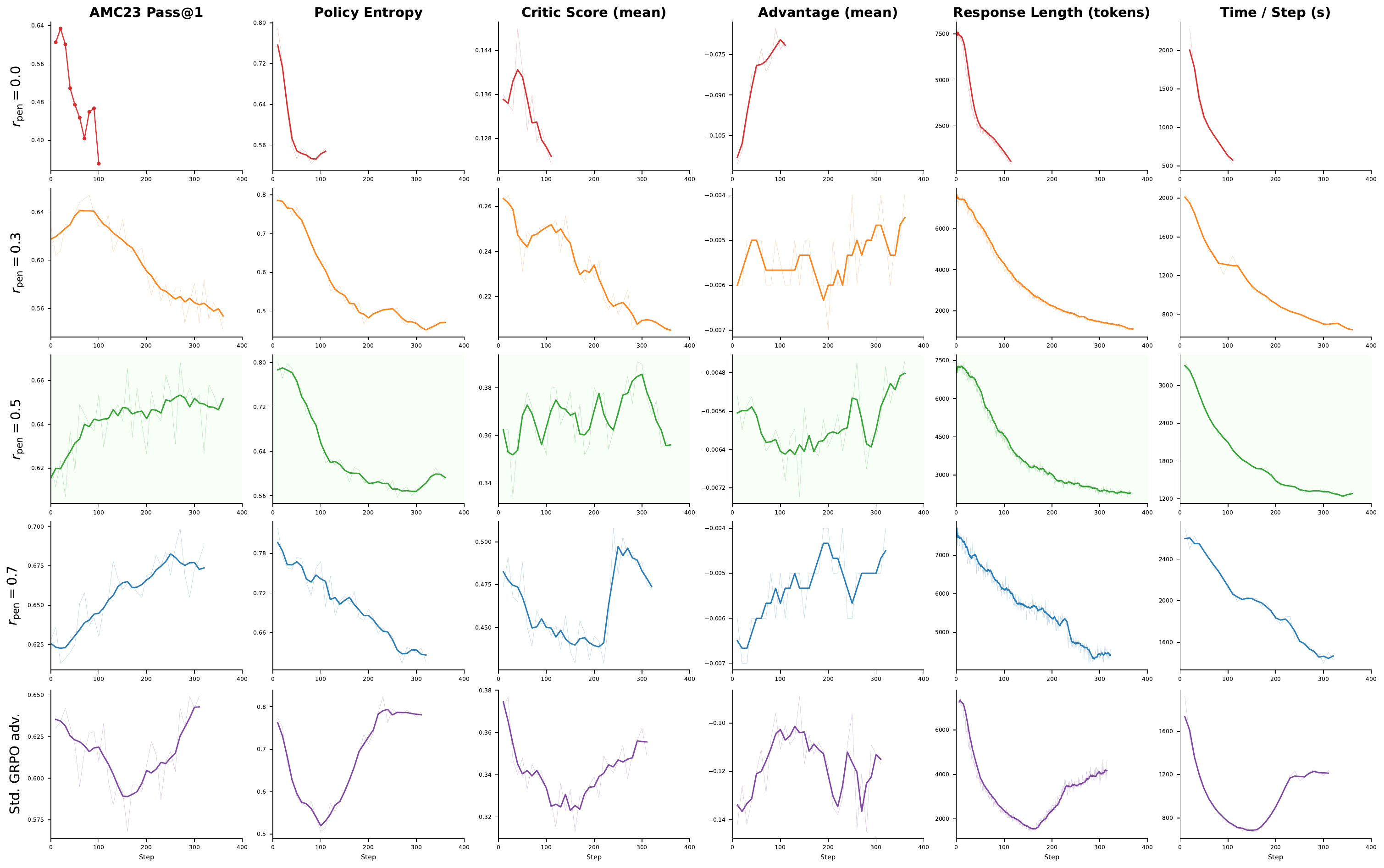}
  \caption{%
    Training curves for ablation experiments on DeepSeek-R1-Distill-Qwen-1.5B.
    \textbf{Rows (top to bottom):} four settings of the discounted reward
    $r_{\text{pen}} \in \{0.0,\,0.3,\,0.5,\,0.7\}$ (red, orange, green, blue), followed by
    a standard GRPO advantage baseline with $r_{\text{pen}}=0.5$ (purple).
    \textbf{Columns (left to right):} (1)~validation AMC23 pass@1 (sampled every 10 steps);
    (2)~policy entropy; (3)~mean critic score; (4)~mean advantage;
    (5)~mean response length (tokens); (6)~wall-clock time per step (seconds).
    For dense per-step metrics (columns 2--6), thin lines show raw values and thick lines
    show smoothed trends.
    The $r_{\text{pen}} = 0.5$ row (green, highlighted) is the full \ourMethod{} configuration
    and achieves the best accuracy--efficiency trade-off.
    Note that $r_{\text{pen}} = 0.3$ starts from step~100 and $r_{\text{pen}} = 0.7$ from
    step~230, as both runs were resumed from earlier checkpoints.%
  }
  \label{fig:ablation_curves:app}
\end{figure}

Figure~\ref{fig:ablation_curves:app} presents the full training trajectories for all four
$r_{\text{pen}}$ ablation settings ($r_{\text{pen}} \in \{0.0,\,0.3,\,0.5,\,0.7\}$ at
fixed $\alpha = 0.1$) together with a standard GRPO advantage baseline
($r_{\text{pen}}=0.5$, standard group-normalised advantage), across six metrics:
validation AMC23 pass@1, policy entropy, mean critic score, mean advantage,
mean response length, and wall-clock time per step.

The most striking contrast is visible in the \textbf{response length} column. Under
$r_{\text{pen}} = 0.0$, the response length collapses from roughly 7{,}600 tokens at step~1
to below 500 tokens by step~100, while AMC23 pass@1 simultaneously plummets from $\sim$55\%
to below 35\%. This confirms that assigning zero reward to incorrect responses removes the
incentive for the model to reason at all: once the policy discovers that short, incorrect
outputs cost nothing, it collapses into near-trivial generations.
In contrast, $r_{\text{pen}} = 0.5$ (our full \ourMethod{} setting) shows a smooth, controlled
compression: response length decreases from $\sim$7{,}300 to $\sim$2{,}600 tokens over 364
steps while AMC23 pass@1 improves steadily to above 65\%.
The $r_{\text{pen}} = 0.3$ setting occupies an intermediate regime---length reduces to
$\sim$2{,}000--3{,}000 tokens but accuracy stabilises around 55\% and remains consistently
lower than the $r_{\text{pen}} = 0.5$ setting, indicating that insufficient penalty still
impairs the incentive structure. The $r_{\text{pen}} = 0.7$ trajectory (steps 230--322)
shows the highest accuracy among all $r_{\text{pen}}$ settings but the least length
reduction, suggesting that an overly generous penalty weakens the compression pressure.

The \textbf{policy entropy} column corroborates these observations: $r_{\text{pen}} = 0.0$
exhibits a rapid entropy collapse consistent with mode collapse, while
$r_{\text{pen}} \in \{0.3, 0.5, 0.7\}$ maintain healthy entropy levels throughout training.
The \textbf{advantage} column mirrors the reward structure: under $r_{\text{pen}} = 0.0$
advantages quickly approach zero (the policy finds a trivial fixed point), whereas
$r_{\text{pen}} = 0.5$ shows consistently non-trivial advantage signals throughout.

The \textbf{standard GRPO advantage} row (purple, bottom) uses identical hyperparameters to
$r_{\text{pen}} = 0.5$ but replaces the ExpThink advantage with standard GRPO
group-normalised advantage. Despite reaching comparable AMC23 accuracy ($\sim$63\%),
its response length converges to $\sim$4{,}600 tokens---nearly double that of \ourMethod{}
($\sim$2{,}600 tokens)---demonstrating that the ExpThink advantage formulation, rather than
the discounted reward alone, is the key driver of efficient compression.

\subsection{Experience Buffer Analysis}
\label{app:buffer}

\begin{wrapfigure}{r}{0.40\linewidth}
  \vspace{-4pt}
  \centering
  \includegraphics[width=\linewidth]{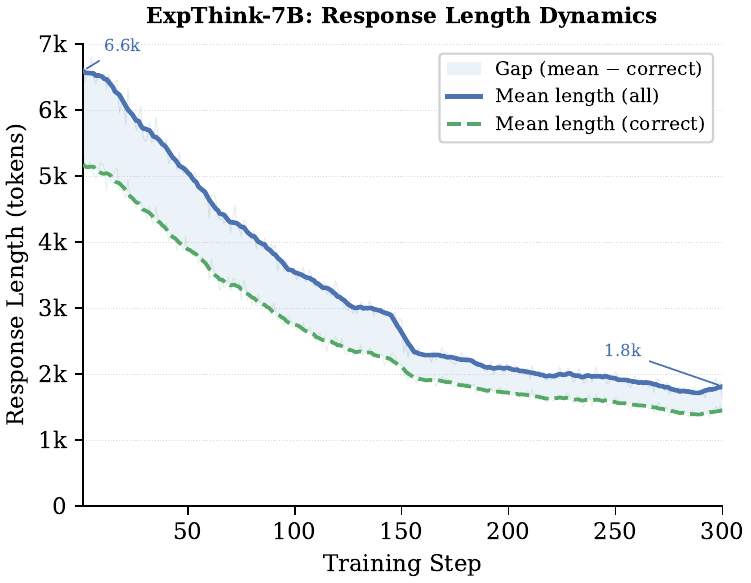}
  \caption{Response length dynamics of \ourMethod{} on DeepSeek-R1-Distill-Qwen-7B.}
  \label{fig:training_trajectory}
  \vspace{-6pt}
\end{wrapfigure}

We analyze the evolution of the training dynamics over 300 update steps using the DeepSeek-R1-Distill-Qwen-7B backbone, tracking the mean response length at each step.

Figure~\ref{fig:training_trajectory} visualises the full training trajectory.
The mean response length drops from approximately 6.6k tokens at step~1 to around 1.8k tokens by step~300—a reduction of roughly 73\%—confirming that the experience-guided reward progressively tightens the length budget as the model accumulates shorter correct solutions.
The mean length of \emph{correct} responses (dashed green) tracks consistently below the overall mean; the shaded region between the two curves represents the \emph{compression gap}, i.e.\ the extra tokens spent by incorrect responses.
Both curves converge toward 1.5--2k tokens in the later training phase, and the gap narrows substantially after step~150, indicating that the model learns to be concise regardless of whether its answer is ultimately correct.

As illustrated in Figure~3 of the main paper, the experience buffer creates an implicit self-evolving curriculum. The running minimum correct-solution length $\ell_q^*$ for each training problem decreases monotonically over training steps, while the average correct-solution length $\bar{\ell}_q$ at each step remains substantially higher---this ``compression gap'' confirms that the model is capable of shorter solutions but only intermittently achieves them without the buffer's guidance. Easy problems (as measured by pass rate) tend to reach their minimum $\ell_q^*$ earlier in training, causing the reward threshold for those problems to tighten quickly; hard problems maintain a higher $\ell_q^*$ for longer, preserving a generous budget that allows the model to focus on correctness rather than brevity. This asymmetric tightening schedule emerges naturally from the model's own training trajectory, without any manual curriculum design.

\section{Case Studies}
\label{app:cases}

We present qualitative comparisons between the Vanilla baseline and \ourMethod~across eight representative problems spanning all evaluated benchmarks: five in-domain mathematical reasoning problems and three out-of-distribution tasks. For each case, we display the full problem statement, followed by the reasoning traces from both models. Long traces are abbreviated with \textit{\textbf{[...](omit $N$ tokens)}} where $N$ is the estimated number of omitted tokens. Token counts are shown next to each model label. Among the eight cases, three demonstrate scenarios where both models answer correctly (\raisebox{0pt}{\small\textcolor{csVanillaBlue}{\textbf{Both Correct}}}), illustrating that \ourMethod~achieves equal accuracy at a fraction of the token cost; the remaining five demonstrate scenarios where only \ourMethod~answers correctly (\raisebox{0pt}{\small\textcolor{csExpGreen}{\textbf{\ourMethod~Correct, Vanilla Incorrect}}}), showing that concise reasoning also improves accuracy by eliminating error-inducing overthinking.

\begin{figure}[htbp]
\centering
\includegraphics[width=\linewidth]{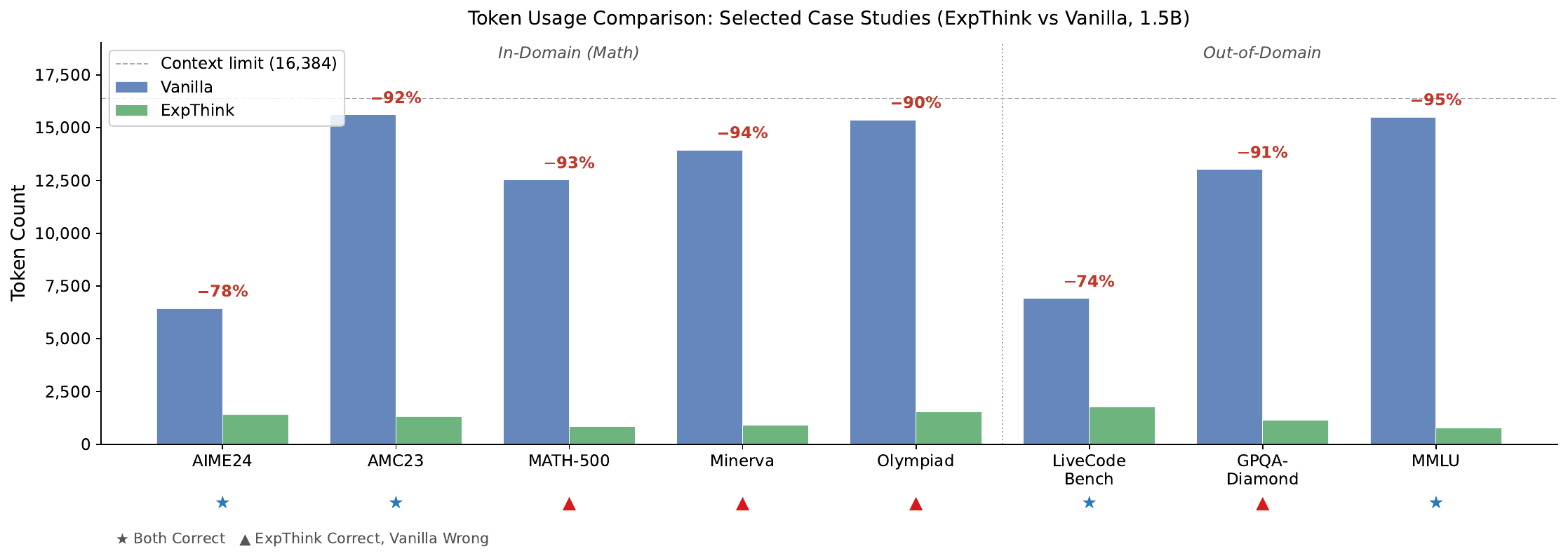}
\caption{Token usage comparison between Vanilla and \ourMethod~(1.5B) on the eight selected case study examples. Red annotations show the relative token reduction. Stars ({\small\textcolor{csVanillaBlue}{$\bigstar$}}) indicate both-correct cases; triangles ({\small\textcolor{csExpGreen}{$\blacktriangle$}}) indicate \ourMethod-only-correct cases. The dashed line marks the 16,384-token context limit.}
\label{fig:case_token_savings}
\end{figure}

\subsection{In-Domain Mathematical Reasoning Cases}
\label{app:cases_indomain}

We select one problem from each of the five in-domain math benchmarks. Two cases (AIME24 and AMC23) show that when both models answer correctly, \ourMethod~uses dramatically fewer tokens---up to 92\% fewer---demonstrating efficient compression without quality loss. Three cases (MATH-500, Minerva, OlympiadBench) show that \ourMethod~answers correctly while Vanilla fails, illustrating how the experience-guided curriculum eliminates the harmful overthinking loops that derail the Vanilla model.

\begin{figure}[p]
\begin{tcolorbox}[
  enhanced,
  colback=white, colframe=black!55, boxrule=0.7pt,
  left=5pt, right=5pt, top=4pt, bottom=4pt,
  title={\textbf{Case Study 1: AIME24} \hfill \small\textit{Both Correct}},
  fonttitle=\small\bfseries,
  attach boxed title to top left={yshift=-2mm, xshift=4mm},
  boxed title style={colback=black!55, colframe=black!55}
]

\noindent\textbf{$\blacktriangleright$~Problem:}
\begin{mdframed}[backgroundcolor=csProblemBg, linewidth=0.4pt, innerleftmargin=6pt, innerrightmargin=6pt, innertopmargin=4pt, innerbottommargin=4pt]
\small
Alice chooses a set $A$ of positive integers. Then Bob lists all finite nonempty sets $B$ of positive integers with the property that the maximum element of $B$ belongs to $A$. Bob's list has 2024 sets. Find the sum of the elements of $A$.
\end{mdframed}

\vspace{4pt}
\noindent\textbf{\textcolor{csVanillaBlue}{\textbullet}~Vanilla} \hfill \texttt{(6,421~tokens)}
\begin{mdframed}[backgroundcolor=csVanillaBg, linewidth=0.3pt, innerleftmargin=6pt, innerrightmargin=6pt, innertopmargin=4pt, innerbottommargin=4pt]
\small
\texttt{<think>}\\
Alright, so I've got this problem where Alice chooses a set $A$ of positive integers, and Bob lists all finite nonempty sets $B$ with the property that the maximum element of $B$ belongs to $A$. Bob's list has 2024 sets. I need to find the sum of elements of $A$. Let me break this down. For each element $a \in A$, the number of valid sets $B$ with maximum element $a$ is $2^{a-1}$, since every subset of $\{1,\ldots,a-1\}$ can be appended to $\{a\}$. So the total count satisfies $\sum_{a \in A} 2^{a-1} = 2024$\ldots

\textit{\textbf{[...](omit 3,425 tokens)}}

$= 2^{10} + 2^9 + 2^8 + 2^7 + 2^6 + 2^5 + 2^3$. This corresponds to exponents $10,9,8,7,6,5,3$, so $A = \{11,10,9,8,7,6,4\}$. Summing: $4+6+7+8+9+10+11 = 55$. Thus the sum of the elements of $A$ is $\boxed{55}$.
\texttt{</think>}

\noindent\textbf{Ground Truth:}~$55$ \quad \textcolor{green!50!black}{$\checkmark$~Correct}
\end{mdframed}

\vspace{4pt}
\noindent\textbf{\textcolor{csExpGreen}{\textbullet}~\ourMethod} \hfill \texttt{(1,416~tokens, \textbf{$-$77.9\% fewer})}
\begin{mdframed}[backgroundcolor=csExpBg, linewidth=0.3pt, innerleftmargin=6pt, innerrightmargin=6pt, innertopmargin=4pt, innerbottommargin=4pt]
\small
\texttt{<think>}\\
Okay, Alice picks set $A$. Bob lists all finite nonempty $B$ where $\max(B)\in A$. For each $a\in A$, subsets of $\{1,\ldots,a\}$ with maximum $a$ number $2^{a-1}$. Total: $\sum_{a\in A} 2^{a-1} = 2024$. Converting 2024 to binary: $11111101000_2 = 2^{10}+2^9+2^8+2^7+2^6+2^5+2^3$. So $A=\{4,6,7,8,9,10,11\}$ and the sum is $4+6+7+8+9+10+11=\boxed{55}$.
\texttt{</think>}

\noindent\textbf{Ground Truth:}~$55$ \quad \textcolor{green!50!black}{$\checkmark$~Correct}
\end{mdframed}

\noindent\textit{The Vanilla model uses 6,421 tokens with extended step-by-step elaboration. \ourMethod~identifies the binary representation key insight directly and solves the problem in 1,416 tokens ($-77.9\%$).}
\end{tcolorbox}
\caption{Case Study 1 (AIME24, Both Correct): \ourMethod{} solves the problem in 1,416 tokens vs.\ Vanilla's 6,421 tokens ($-77.9\%$).}
\label{fig:case1_aime24}
\end{figure}

\begin{figure}[p]
\begin{tcolorbox}[
  enhanced,
  colback=white, colframe=black!55, boxrule=0.7pt,
  left=5pt, right=5pt, top=4pt, bottom=4pt,
  title={\textbf{Case Study 2: AMC23} \hfill \small\textit{Both Correct}},
  fonttitle=\small\bfseries,
  attach boxed title to top left={yshift=-2mm, xshift=4mm},
  boxed title style={colback=black!55, colframe=black!55}
]

\noindent\textbf{$\blacktriangleright$~Problem:}
\begin{mdframed}[backgroundcolor=csProblemBg, linewidth=0.4pt, innerleftmargin=6pt, innerrightmargin=6pt, innertopmargin=4pt, innerbottommargin=4pt]
\small
When the roots of the polynomial $P(x) = (x-1)^1(x-2)^2(x-3)^3\cdots(x-10)^{10}$ are removed from the number line, what remains is the union of 11 disjoint open intervals. On how many of these intervals is $P(x)$ positive?
\end{mdframed}

\vspace{4pt}
\noindent\textbf{\textcolor{csVanillaBlue}{\textbullet}~Vanilla} \hfill \texttt{(15,606~tokens)}
\begin{mdframed}[backgroundcolor=csVanillaBg, linewidth=0.3pt, innerleftmargin=6pt, innerrightmargin=6pt, innertopmargin=4pt, innerbottommargin=4pt]
\small
\texttt{<think>}\\
Okay, so I have this polynomial $P(x) = (x-1)^1(x-2)^2\cdots(x-10)^{10}$. The roots are $1,2,\ldots,10$ with multiplicities $1,2,\ldots,10$. I need to determine on how many of the 11 resulting intervals $P(x) > 0$. Let me analyze the sign changes at each root based on multiplicity\ldots

\textit{\textbf{[...](omit 10,314 tokens)}}

The positive intervals are $(10,\infty),(8,9),(6,7),(4,5),(2,3),(1,2)$. Thus the polynomial $P(x)$ is positive on $\boxed{6}$ disjoint open intervals.
\texttt{</think>}

\noindent\textbf{Ground Truth:}~$6$ \quad \textcolor{green!50!black}{$\checkmark$~Correct}
\end{mdframed}

\vspace{4pt}
\noindent\textbf{\textcolor{csExpGreen}{\textbullet}~\ourMethod} \hfill \texttt{(1,309~tokens, \textbf{$-$91.6\% fewer})}
\begin{mdframed}[backgroundcolor=csExpBg, linewidth=0.3pt, innerleftmargin=6pt, innerrightmargin=6pt, innertopmargin=4pt, innerbottommargin=4pt]
\small
\texttt{<think>}\\
The sign changes at each root according to its multiplicity. Even-multiplicity roots do not change sign; odd-multiplicity roots do. For $x > 10$, $P(x)>0$. Working leftward, the sign flips only at odd-multiplicity roots ($k=1,3,5,7,9$). The intervals where $P(x)>0$ are $(-\infty,1),(2,3),(4,5),(6,7),(8,9),(10,\infty)$. That is $\boxed{6}$ intervals.
\texttt{</think>}

\noindent\textbf{Ground Truth:}~$6$ \quad \textcolor{green!50!black}{$\checkmark$~Correct}
\end{mdframed}

\noindent\textit{The Vanilla model uses 15,606 tokens (nearly at the context limit) cycling through exhaustive case analysis. \ourMethod~recognizes the parity key immediately and solves the problem in 1,309 tokens ($-91.6\%$).}
\end{tcolorbox}
\caption{Case Study 2 (AMC23, Both Correct): \ourMethod{} solves the problem in 1,309 tokens vs.\ Vanilla's 15,606 tokens ($-91.6\%$).}
\label{fig:case2_amc23}
\end{figure}

\begin{figure}[p]
\begin{tcolorbox}[
  enhanced,
  colback=white, colframe=black!55, boxrule=0.7pt,
  left=5pt, right=5pt, top=4pt, bottom=4pt,
  title={\textbf{Case Study 3: MATH-500} \hfill \small\textit{\ourMethod~Correct, Vanilla Incorrect}},
  fonttitle=\small\bfseries,
  attach boxed title to top left={yshift=-2mm, xshift=4mm},
  boxed title style={colback=black!55, colframe=black!55}
]

\noindent\textbf{$\blacktriangleright$~Problem:}
\begin{mdframed}[backgroundcolor=csProblemBg, linewidth=0.4pt, innerleftmargin=6pt, innerrightmargin=6pt, innertopmargin=4pt, innerbottommargin=4pt]
\small
Each week, between 30 and 50 students show up for an archery class run by Betty and Wilma. Usually the students break up into groups of equal size for target practice. However, this week, Betty noticed that she could not break the students up into multiple groups of equal size. Wilma noticed that if she and Betty both joined the students in practicing, they still could not break the archers up into groups of equal size. How many students showed up to the archery class this week?
\end{mdframed}

\vspace{4pt}
\noindent\textbf{\textcolor{csVanillaBlue}{\textbullet}~Vanilla} \hfill \texttt{(12,523~tokens)}
\begin{mdframed}[backgroundcolor=csVanillaBg, linewidth=0.3pt, innerleftmargin=6pt, innerrightmargin=6pt, innertopmargin=4pt, innerbottommargin=4pt]
\small
\texttt{<think>}\\
Okay, so we need $N \in [30,50]$ such that $N$ is prime and $N+2$ is also prime. Betty can't split $N$ students into equal groups $\Rightarrow$ $N$ is prime. Adding Betty and Wilma gives $N+2$, also prime. Checking twin primes in $[30,50]$: $(31,33)$, $(37,39)$, $(41,43)$, $(43,45)$, $(47,49)$\ldots Let me check each\ldots Wait, $N+1$ or $N+2$? Let me reread\ldots

\textit{\textbf{[...](omit 9,694 tokens)}}

After exhaustive checking, the number 44 is the minimal number that fits all conditions. $\boxed{44}$
\texttt{</think>}

\noindent\textbf{Ground Truth:}~$41$ \quad \textcolor{red!70!black}{$\times$~Wrong}
\end{mdframed}

\vspace{4pt}
\noindent\textbf{\textcolor{csExpGreen}{\textbullet}~\ourMethod} \hfill \texttt{(829~tokens, \textbf{$-$93.4\% fewer})}
\begin{mdframed}[backgroundcolor=csExpBg, linewidth=0.3pt, innerleftmargin=6pt, innerrightmargin=6pt, innertopmargin=4pt, innerbottommargin=4pt]
\small
\texttt{<think>}\\
$N \in [30,50]$ must be prime (Betty cannot split into equal groups). Adding Betty and Wilma gives $N+2$ which must also be prime. Checking: $31+2=33$ (not prime); $37+2=39$ (not prime); $41+2=43$ (prime!). So $N=41$. $\boxed{41}$
\texttt{</think>}

\noindent\textbf{Ground Truth:}~$41$ \quad \textcolor{green!50!black}{$\checkmark$~Correct}
\end{mdframed}

\noindent\textit{The Vanilla model uses 12,523 tokens, loses track of the correct constraint ($N+2$ must be prime), and returns the wrong answer 44. \ourMethod~correctly applies twin-prime reasoning in 829 tokens ($-93.4\%$).}
\end{tcolorbox}
\caption{Case Study 3 (MATH-500, \ourMethod{} Correct / Vanilla Incorrect): \ourMethod{} applies twin-prime reasoning in 829 tokens; Vanilla uses 12,523 tokens and returns the wrong answer.}
\label{fig:case3_math500}
\end{figure}

\begin{figure}[p]
\begin{tcolorbox}[
  enhanced,
  colback=white, colframe=black!55, boxrule=0.7pt,
  left=5pt, right=5pt, top=4pt, bottom=4pt,
  title={\textbf{Case Study 4: Minerva Math} \hfill \small\textit{\ourMethod~Correct, Vanilla Incorrect}},
  fonttitle=\small\bfseries,
  attach boxed title to top left={yshift=-2mm, xshift=4mm},
  boxed title style={colback=black!55, colframe=black!55}
]

\noindent\textbf{$\blacktriangleright$~Problem:}
\begin{mdframed}[backgroundcolor=csProblemBg, linewidth=0.4pt, innerleftmargin=6pt, innerrightmargin=6pt, innertopmargin=4pt, innerbottommargin=4pt]
\small
If the Sun's absolute magnitude is $+5$, find the luminosity of a star of magnitude $0$ in ergs/s. A useful constant: the luminosity of the Sun is $3.83 \times 10^{33}$ ergs/s.
\end{mdframed}

\vspace{4pt}
\noindent\textbf{\textcolor{csVanillaBlue}{\textbullet}~Vanilla} \hfill \texttt{(13,942~tokens)}
\begin{mdframed}[backgroundcolor=csVanillaBg, linewidth=0.3pt, innerleftmargin=6pt, innerrightmargin=6pt, innertopmargin=4pt, innerbottommargin=4pt]
\small
\texttt{<think>}\\
The magnitude difference is $\Delta M = 5 - 0 = 5$. The flux ratio is $F_2/F_1 = 10^{0.4\Delta M} = 10^2 = 100$. But wait, a higher magnitude means dimmer, so the star at magnitude 0 is brighter than the Sun at magnitude 5. The luminosity ratio is $L_\star / L_\odot = 100$. But let me double-check the sign convention\ldots

\textit{\textbf{[...](omit 10,532 tokens)}}

Actually, using $M_1 - M_2 = -2.5\log(L_2/L_1)$: $(5-0) = 5 = -2.5\log(L/L_\odot)$, so $\log(L/L_\odot) = -2$, giving $L = 0.1 L_\odot = 3.83\times10^{32}$ ergs/s. $\boxed{3.83\times10^{32}~\text{ergs/s}}$
\texttt{</think>}

\noindent\textbf{Ground Truth:}~$3.83\times10^{35}$ ergs/s \quad \textcolor{red!70!black}{$\times$~Wrong}
\end{mdframed}

\vspace{4pt}
\noindent\textbf{\textcolor{csExpGreen}{\textbullet}~\ourMethod} \hfill \texttt{(892~tokens, \textbf{$-$93.6\% fewer})}
\begin{mdframed}[backgroundcolor=csExpBg, linewidth=0.3pt, innerleftmargin=6pt, innerrightmargin=6pt, innertopmargin=4pt, innerbottommargin=4pt]
\small
\texttt{<think>}\\
A star at magnitude 0 is brighter than the Sun at magnitude $+5$. Using $L_2/L_1 = 10^{0.4(M_1-M_2)}$: $L_\star/L_\odot = 10^{0.4\times5} = 10^2 = 100$. So $L_\star = 100\times3.83\times10^{33} = 3.83\times10^{35}$ ergs/s. $\boxed{3.83\times10^{35}}$
\texttt{</think>}

\noindent\textbf{Ground Truth:}~$3.83\times10^{35}$ ergs/s \quad \textcolor{green!50!black}{$\checkmark$~Correct}
\end{mdframed}

\noindent\textit{The Vanilla model spends 13,942 tokens debating sign conventions, flips the ratio, and returns $3.83\times10^{32}$ ergs/s (wrong by a factor of $10^3$). \ourMethod~applies the magnitude formula directly in 892 tokens ($-93.6\%$) and reaches the correct answer.}
\end{tcolorbox}
\caption{Case Study 4 (Minerva Math, \ourMethod{} Correct / Vanilla Incorrect): \ourMethod{} applies the magnitude formula in 892 tokens; Vanilla uses 13,942 tokens and returns an answer wrong by $10^3$.}
\label{fig:case4_minerva}
\end{figure}

\begin{figure}[p]
\begin{tcolorbox}[
  enhanced,
  colback=white, colframe=black!55, boxrule=0.7pt,
  left=5pt, right=5pt, top=4pt, bottom=4pt,
  title={\textbf{Case Study 5: OlympiadBench} \hfill \small\textit{\ourMethod~Correct, Vanilla Incorrect}},
  fonttitle=\small\bfseries,
  attach boxed title to top left={yshift=-2mm, xshift=4mm},
  boxed title style={colback=black!55, colframe=black!55}
]

\noindent\textbf{$\blacktriangleright$~Problem:}
\begin{mdframed}[backgroundcolor=csProblemBg, linewidth=0.4pt, innerleftmargin=6pt, innerrightmargin=6pt, innertopmargin=4pt, innerbottommargin=4pt]
\small
Let $EUCLID$ be a hexagon inscribed in a circle of radius 5. Given that $EU=UC=LI=ID=6$ and $CL=DE$, compute $CL$.
\end{mdframed}

\vspace{4pt}
\noindent\textbf{\textcolor{csVanillaBlue}{\textbullet}~Vanilla} \hfill \texttt{(15,361~tokens)}
\begin{mdframed}[backgroundcolor=csVanillaBg, linewidth=0.3pt, innerleftmargin=6pt, innerrightmargin=6pt, innertopmargin=4pt, innerbottommargin=4pt]
\small
\texttt{<think>}\\
For a chord of length $\ell$ in a circle of radius $R$, the central angle satisfies $\ell = 2R\sin(\theta/2)$. For $EU=UC=LI=ID=6$ with $R=5$: $6 = 10\sin\alpha$, so $\sin\alpha = 0.6$ and $\alpha = \arcsin(0.6)$. The six central angles sum to $2\pi$: $4\alpha + 2\gamma = 2\pi$ where $\gamma$ is the central angle for $CL$ and $DE$. Hmm, but maybe the hexagon is regular except for these two sides\ldots

\textit{\textbf{[...](omit 9,144 tokens)}}

Since the hexagon distributes the central angles evenly, each central angle is $2\pi/6 = \pi/3$. The chord $CL$ subtends $\pi/3$, giving $CL = 2\times5\times\sin(\pi/6)=6$. $\boxed{6}$
\texttt{</think>}

\noindent\textbf{Ground Truth:}~$\dfrac{14}{5}$ \quad \textcolor{red!70!black}{$\times$~Wrong}
\end{mdframed}

\vspace{4pt}
\noindent\textbf{\textcolor{csExpGreen}{\textbullet}~\ourMethod} \hfill \texttt{(1,525~tokens, \textbf{$-$90.1\% fewer})}
\begin{mdframed}[backgroundcolor=csExpBg, linewidth=0.3pt, innerleftmargin=6pt, innerrightmargin=6pt, innertopmargin=4pt, innerbottommargin=4pt]
\small
\texttt{<think>}\\
For $EU=UC=LI=ID=6$ in a circle of radius 5, each subtends central angle $\alpha$ where $\sin(\alpha/2)=3/5=0.6$, so $\cos(\alpha/2)=4/5$. Total angles: $4\alpha+2\gamma=2\pi$, giving $\gamma=\pi-2\alpha$. Then $\sin(\gamma/2)=\sin(\pi/2-\alpha)=\cos\alpha=1-2\sin^2(\alpha/2)=1-2(9/25)=7/25$. So $CL=2\times5\times(7/25)=\boxed{14/5}$.
\texttt{</think>}

\noindent\textbf{Ground Truth:}~$\dfrac{14}{5}$ \quad \textcolor{green!50!black}{$\checkmark$~Correct}
\end{mdframed}

\noindent\textit{The Vanilla model incorrectly assumes the hexagon must be regular (all sides equal), returns $CL=6$, and uses 15,361 tokens. \ourMethod~applies the central angle constraint correctly in 1,525 tokens ($-90.1\%$).}
\end{tcolorbox}
\caption{Case Study 5 (OlympiadBench, \ourMethod{} Correct / Vanilla Incorrect): \ourMethod{} correctly solves the hexagon problem in 1,525 tokens; Vanilla uses 15,361 tokens and returns the wrong answer.}
\label{fig:case5_olympiad}
\end{figure}

\clearpage

\subsection{Out-of-Distribution Generalization Cases}
\label{app:cases_ood}

We select one problem from each of the three OOD benchmarks. These tasks were never seen during training, yet \ourMethod~generalizes its efficient reasoning style to code generation (LiveCodeBench), graduate-level science (GPQA-Diamond), and multi-domain knowledge (MMLU). Notably, on MMLU, \ourMethod~achieves a 95\% token reduction while maintaining correctness, and on GPQA-Diamond, \ourMethod~succeeds where the verbose Vanilla trace leads to an incorrect conclusion.

\begin{figure}[p]
\begin{tcolorbox}[
  enhanced,
  colback=white, colframe=black!55, boxrule=0.7pt,
  left=5pt, right=5pt, top=4pt, bottom=4pt,
  title={\textbf{Case Study 6: MMLU} \hfill \small\textit{Both Correct}},
  fonttitle=\small\bfseries,
  attach boxed title to top left={yshift=-2mm, xshift=4mm},
  boxed title style={colback=black!55, colframe=black!55}
]

\noindent\textbf{$\blacktriangleright$~Problem:}
\begin{mdframed}[backgroundcolor=csProblemBg, linewidth=0.4pt, innerleftmargin=6pt, innerrightmargin=6pt, innertopmargin=4pt, innerbottommargin=4pt]
\small
The rotational correlation time of a spherical object of radius $r$ immersed in a liquid with viscosity $\eta$ is given by the Stokes equation: $\tau_c = 4\pi\eta r^3/(3k_BT)$. A small molecule with relative molecular mass $M_r = 100$ in water at 298 K has $\tau_c = 50$ ps. Assuming that $M_r$ is proportional to the molecular volume, estimate $\tau_c$ for a small protein with $M_r = 8400$.\\
A)~420 ns. \quad B)~42 ns. \quad C)~4.2 ns. \quad D)~420 ps.
\end{mdframed}

\vspace{4pt}
\noindent\textbf{\textcolor{csVanillaBlue}{\textbullet}~Vanilla} \hfill \texttt{(15,493~tokens)}
\begin{mdframed}[backgroundcolor=csVanillaBg, linewidth=0.3pt, innerleftmargin=6pt, innerrightmargin=6pt, innertopmargin=4pt, innerbottommargin=4pt]
\small
\texttt{<think>}\\
$\tau_c \propto r^3$ and $M_r \propto V \propto r^3$, so $\tau_c \propto M_r$. Then $\tau_{c2}/\tau_{c1} = M_{r2}/M_{r1} = 84$, giving $\tau_{c2} = 50\times84 = 4200$ ps $= 4.2$ ns\ldots But wait, is $\tau_c \propto r^3$ or $\propto r^6$? Let me recheck the Stokes equation\ldots

\textit{\textbf{[...](omit 9,682 tokens)}}

After reconsidering, the scaling factor when $M_r\propto V$ gives $\tau_{c2}=50\times84 = 4200$ ps $= 4.2$ ns. $\boxed{C}$
\texttt{</think>}

\noindent\textbf{Ground Truth:}~C \quad \textcolor{green!50!black}{$\checkmark$~Correct}
\end{mdframed}

\vspace{4pt}
\noindent\textbf{\textcolor{csExpGreen}{\textbullet}~\ourMethod} \hfill \texttt{(768~tokens, \textbf{$-$95.0\% fewer})}
\begin{mdframed}[backgroundcolor=csExpBg, linewidth=0.3pt, innerleftmargin=6pt, innerrightmargin=6pt, innertopmargin=4pt, innerbottommargin=4pt]
\small
\texttt{<think>}\\
Since $M_r \propto V \propto r^3$ and $\tau_c \propto r^3$, we have $\tau_c \propto M_r$. Thus $\tau_{c2} = \tau_{c1}\times(M_{r2}/M_{r1}) = 50\text{ ps}\times(8400/100) = 50\times84 = 4200\text{ ps} = 4.2\text{ ns}$. $\boxed{C}$
\texttt{</think>}

\noindent\textbf{Ground Truth:}~C \quad \textcolor{green!50!black}{$\checkmark$~Correct}
\end{mdframed}

\noindent\textit{The Vanilla model uses 15,493 tokens, repeatedly second-guessing the scaling relationship before arriving at the correct answer. \ourMethod~applies the proportionality argument immediately in 768 tokens ($-95.0\%$).}
\end{tcolorbox}
\caption{Case Study 6 (MMLU, Both Correct): \ourMethod{} answers in 768 tokens vs.\ Vanilla's 15,493 tokens ($-95.0\%$).}
\label{fig:case6_mmlu}
\end{figure}

\begin{figure}[p]
\begin{tcolorbox}[
  enhanced,
  colback=white, colframe=black!55, boxrule=0.7pt,
  left=5pt, right=5pt, top=4pt, bottom=4pt,
  title={\textbf{Case Study 7: GPQA-Diamond} \hfill \small\textit{\ourMethod~Correct, Vanilla Incorrect}},
  fonttitle=\small\bfseries,
  attach boxed title to top left={yshift=-2mm, xshift=4mm},
  boxed title style={colback=black!55, colframe=black!55}
]

\noindent\textbf{$\blacktriangleright$~Problem:}
\begin{mdframed}[backgroundcolor=csProblemBg, linewidth=0.4pt, innerleftmargin=6pt, innerrightmargin=6pt, innertopmargin=4pt, innerbottommargin=4pt]
\small
What is the energy of the Relativistic Heavy Ion Collider (RHIC) so that the speed of the nucleus $X$ is equal to $0.96c$? Knowing that $X$ is defined as Li with 3 neutrons. (Precision: $10^{-4}$.)\\
A)~20.132 GeV. \quad B)~21.419 GeV. \quad C)~23.069 GeV. \quad D)~18.475 GeV.
\end{mdframed}

\vspace{4pt}
\noindent\textbf{\textcolor{csVanillaBlue}{\textbullet}~Vanilla} \hfill \texttt{(13,024~tokens)}
\begin{mdframed}[backgroundcolor=csVanillaBg, linewidth=0.3pt, innerleftmargin=6pt, innerrightmargin=6pt, innertopmargin=4pt, innerbottommargin=4pt]
\small
\texttt{<think>}\\
Li with 3 neutrons: atomic number 3, mass number $3+3=6$. Lorentz factor: $\gamma = 1/\sqrt{1-0.96^2} = 1/\sqrt{0.0784} = 25/7\approx3.571$. Rest mass of Li-6: $m = 6\times938.3 = 5629.8$ MeV$/c^2$. Total energy: $E = \gamma mc^2 = (25/7)\times5629.8 = 20106.4$ MeV$\approx20.1$ GeV. Hmm, let me recalculate the mass more carefully using atomic mass units\ldots

\textit{\textbf{[...](omit 6,929 tokens)}}

Using atomic masses more carefully: $m = 6\times940.1\approx5640.6$ MeV. $E = (25/7)\times6.465 = 23.09$ GeV. The closest option is C) 23.069 GeV. $\boxed{C}$
\texttt{</think>}

\noindent\textbf{Ground Truth:}~A \quad \textcolor{red!70!black}{$\times$~Wrong}
\end{mdframed}

\vspace{4pt}
\noindent\textbf{\textcolor{csExpGreen}{\textbullet}~\ourMethod} \hfill \texttt{(1,143~tokens, \textbf{$-$91.2\% fewer})}
\begin{mdframed}[backgroundcolor=csExpBg, linewidth=0.3pt, innerleftmargin=6pt, innerrightmargin=6pt, innertopmargin=4pt, innerbottommargin=4pt]
\small
\texttt{<think>}\\
Li with 3 neutrons: 3 protons + 3 neutrons, mass number 6. Rest mass: $m \approx 6\times938$ MeV/$c^2 = 5628$ MeV/$c^2 = 5.628$ GeV/$c^2$. Lorentz factor at $v=0.96c$: $\gamma = 1/\sqrt{1-0.9216} = 1/0.28 \approx 3.5714$. Total energy: $E = \gamma mc^2 = 3.5714\times5.628 \approx 20.1$ GeV. Closest option: A) 20.132 GeV. $\boxed{A}$
\texttt{</think>}

\noindent\textbf{Ground Truth:}~A \quad \textcolor{green!50!black}{$\checkmark$~Correct}
\end{mdframed}

\noindent\textit{The Vanilla model uses 13,024 tokens, switches between nucleon mass conventions mid-computation, and returns the wrong answer C (23.069 GeV). \ourMethod~uses standard proton mass ($938$ MeV) consistently and finds the correct answer A (20.132 GeV) in 1,143 tokens ($-91.2\%$).}
\end{tcolorbox}
\caption{Case Study 7 (GPQA-Diamond, \ourMethod{} Correct / Vanilla Incorrect): \ourMethod{} finds the correct answer in 1,143 tokens; Vanilla uses 13,024 tokens and returns the wrong answer.}
\label{fig:case7_gpqa}
\end{figure}

\begin{figure}[p]
\begin{tcolorbox}[
  enhanced,
  colback=white, colframe=black!55, boxrule=0.7pt,
  left=5pt, right=5pt, top=4pt, bottom=4pt,
  title={\textbf{Case Study 8: LiveCodeBench} \hfill \small\textit{Both Correct}},
  fonttitle=\small\bfseries,
  attach boxed title to top left={yshift=-2mm, xshift=4mm},
  boxed title style={colback=black!55, colframe=black!55}
]

\noindent\textbf{$\blacktriangleright$~Problem:}
\begin{mdframed}[backgroundcolor=csProblemBg, linewidth=0.4pt, innerleftmargin=6pt, innerrightmargin=6pt, innertopmargin=4pt, innerbottommargin=4pt]
\small
There are $N$ people labeled 1 to $N$. Person $i$ has integer $A_i$. Among those for whom ``none of the other $N-1$ people has the same integer,'' find the one with the greatest integer and print their label. If no such person exists, print $-1$.
\end{mdframed}

\vspace{4pt}
\noindent\textbf{\textcolor{csVanillaBlue}{\textbullet}~Vanilla} \hfill \texttt{(6,908~tokens)}
\begin{mdframed}[backgroundcolor=csVanillaBg, linewidth=0.3pt, innerleftmargin=6pt, innerrightmargin=6pt, innertopmargin=4pt, innerbottommargin=4pt]
\small
\texttt{<think>}\\
I need to find people with unique $A_i$ values, then return the label of the one with the largest $A_i$. I'll use a frequency counter. If the max unique value maps to multiple labels, wait, no---uniqueness ensures there's exactly one label for each unique value\ldots

\textit{\textbf{[...](omit 6,366 tokens)}}

\texttt{</think>}
\begin{lstlisting}[style=cscode]
from collections import Counter
n = int(input())
A = list(map(int, input().split()))
freq = Counter(A)
unique_A = [a for a in A if freq[a] == 1]
if not unique_A:
    print(-1)
else:
    max_A = max(unique_A)
    print(A.index(max_A) + 1)
\end{lstlisting}
\noindent\textbf{Ground Truth:}~PASS \quad \textcolor{green!50!black}{$\checkmark$~Correct}
\end{mdframed}

\vspace{4pt}
\noindent\textbf{\textcolor{csExpGreen}{\textbullet}~\ourMethod} \hfill \texttt{(1,776~tokens, \textbf{$-$74.3\% fewer})}
\begin{mdframed}[backgroundcolor=csExpBg, linewidth=0.3pt, innerleftmargin=6pt, innerrightmargin=6pt, innertopmargin=4pt, innerbottommargin=4pt]
\small
\texttt{<think>}\\
Build a frequency dict. Collect candidates $(A_i, i{+}1)$ with count 1. If empty, print $-1$; else print the label of the entry with max $A_i$.
\texttt{</think>}
\begin{lstlisting}[style=cscode]
from collections import Counter
n = int(input())
a = list(map(int, input().split()))
freq = Counter(a)
candidates = [(val, idx+1) for idx, val in enumerate(a) if freq[val] == 1]
if not candidates:
    print(-1)
else:
    print(max(candidates)[1])
\end{lstlisting}
\noindent\textbf{Ground Truth:}~PASS \quad \textcolor{green!50!black}{$\checkmark$~Correct}
\end{mdframed}

\noindent\textit{Both models produce correct solutions. \ourMethod~uses 1,776 tokens vs.\ Vanilla's 6,908 tokens ($-74.3\%$), writing more direct code without extended self-verification loops.}
\end{tcolorbox}
\caption{Case Study 8 (LiveCodeBench, Both Correct): \ourMethod{} produces correct code in 1,776 tokens vs.\ Vanilla's 6,908 tokens ($-74.3\%$).}
\label{fig:case8_livecodebench}
\end{figure}

\newpage

\clearpage

\section*{NeurIPS Paper Checklist}

\begin{enumerate}


\item {\bf Claims}
    \item[] Question: Do the main claims made in the abstract and introduction accurately reflect the paper's contributions and scope?
    \item[] Answer: \answerYes{}
    \item[] Justification: The abstract and introduction clearly state the three core contributions:
    (1) identification of a structural limitation in existing RL-based CoT compression methods
    regarding discarded historical trajectory information; (2) the proposed \ourMethod{}
    framework comprising a self-evolving compression curriculum, an experience-guided
    three-tier reward, and difficulty-adaptive advantage estimation; and (3) the experimental
    scope (mathematical reasoning benchmarks with three backbone models) and key quantitative
    claims (up to 77\% response length reduction with consistent accuracy improvement, stable
    training dynamics, and up to $3\times$ higher IPT than the vanilla baseline). All claims
    are substantiated by Table~\ref{tab:main_results}, Table~\ref{tab:ablation} and Table~\ref{tab:ood_results} in
    Section~\ref{sec:experiments}.
    \item[] Guidelines:
    \begin{itemize}
        \item The answer \answerNA{} means that the abstract and introduction do not include the claims made in the paper.
        \item The abstract and/or introduction should clearly state the claims made, including the contributions made in the paper and important assumptions and limitations. A \answerNo{} or \answerNA{} answer to this question will not be perceived well by the reviewers.
        \item The claims made should match theoretical and experimental results, and reflect how much the results can be expected to generalize to other settings.
        \item It is fine to include aspirational goals as motivation as long as it is clear that these goals are not attained by the paper.
    \end{itemize}

\item {\bf Limitations}
    \item[] Question: Does the paper discuss the limitations of the work performed by the authors?
    \item[] Answer: \answerYes{}
    \item[] Justification: Section~5 (Conclusion) contains a dedicated limitations paragraph discussing the following limitations: (1) training and evaluation are restricted to mathematical reasoning with verifiable answers, and transfer to interactive tool use or multimodal settings remains unconfirmed; (2) reliance on reliable answer verification, which may not hold for tasks with ambiguous
    or partially correct outputs; (3) the absence of a controlled comparison against DAPO under matched compute, which is left to future work; and (4) the tolerance margin $\alpha$ and discounted reward $r_{\text{pen}}$ are empirical choices, and a systematic study of $\alpha$ sensitivity across diverse domains is left to future work.
    \item[] Guidelines:
    \begin{itemize}
        \item The answer \answerNA{} means that the paper has no limitation while the answer \answerNo{} means that the paper has limitations, but those are not discussed in the paper.
        \item The authors are encouraged to create a separate ``Limitations'' section in their paper.
        \item The paper should point out any strong assumptions and how robust the results are to violations of these assumptions (e.g., independence assumptions, noiseless settings, model well-specification, asymptotic approximations only holding locally). The authors should reflect on how these assumptions might be violated in practice and what the implications would be.
        \item The authors should reflect on the scope of the claims made, e.g., if the approach was only tested on a few datasets or with a few runs. In general, empirical results often depend on implicit assumptions, which should be articulated.
        \item The authors should reflect on the factors that influence the performance of the approach. For example, a facial recognition algorithm may perform poorly when image resolution is low or images are taken in low lighting. Or a speech-to-text system might not be used reliably to provide closed captions for online lectures because it fails to handle technical jargon.
        \item The authors should discuss the computational efficiency of the proposed algorithms and how they scale with dataset size.
        \item If applicable, the authors should discuss possible limitations of their approach to address problems of privacy and fairness.
        \item While the authors might fear that complete honesty about limitations might be used by reviewers as grounds for rejection, a worse outcome might be that reviewers discover limitations that aren't acknowledged in the paper. The authors should use their best judgment and recognize that individual actions in favor of transparency play an important role in developing norms that preserve the integrity of the community. Reviewers will be specifically instructed to not penalize honesty concerning limitations.
    \end{itemize}

\item {\bf Theory assumptions and proofs}
    \item[] Question: For each theoretical result, does the paper provide the full set of assumptions and a complete (and correct) proof?
    \item[] Answer: \answerNA{}
    \item[] Justification: The paper does not include formal theorems or proofs. The key equations (buffer update rule, three-tier reward, count-based advantage estimator) are presented as algorithm design choices with analytical justification in the text rather than as formal theoretical results.
    \item[] Guidelines:
    \begin{itemize}
        \item The answer \answerNA{} means that the paper does not include theoretical results.
        \item All the theorems, formulas, and proofs in the paper should be numbered and cross-referenced.
        \item All assumptions should be clearly stated or referenced in the statement of any theorems.
        \item The proofs can either appear in the main paper or the supplemental material, but if they appear in the supplemental material, the authors are encouraged to provide a short proof sketch to provide intuition.
        \item Inversely, any informal proof provided in the core of the paper should be complemented by formal proofs provided in appendix or supplemental material.
        \item Theorems and Lemmas that the proof relies upon should be properly referenced.
    \end{itemize}

    \item {\bf Experimental result reproducibility}
    \item[] Question: Does the paper fully disclose all the information needed to reproduce the main
    experimental results of the paper to the extent that it affects the main claims and/or
    conclusions of the paper (regardless of whether the code and data are provided or not)?
    \item[] Answer: \answerYes{}
    \item[] Justification: Appendix~\ref{app:training_params} provides all training hyperparameters
    (hardware: 8$\times$A800 80\,GB; batch size, learning rate, context length, FSDP
    configuration, PPO clip bounds, group size, and all method-specific hyperparameters
    $\alpha$ and $r_{\text{pen}}$). Appendix~\ref{app:datasets} describes all datasets and
    Appendix~\ref{app:eval_protocol} defines the evaluation protocol. The algorithm is fully
    specified in Section~3.
    \item[] Guidelines:
    \begin{itemize}
        \item The answer \answerNA{} means that the paper does not include experiments.
        \item If the paper includes experiments, a \answerNo{} answer to this question will not be perceived well by the reviewers: Making the paper reproducible is important, regardless of whether the code and data are provided or not.
        \item If the contribution is a dataset and\slash or model, the authors should describe the steps taken to make their results reproducible or verifiable.
        \item Depending on the contribution, reproducibility can be accomplished in various ways. For example, if the contribution is a novel architecture, describing the architecture fully might suffice, or if the contribution is a specific model and empirical evaluation, it may be necessary to either make it possible for others to replicate the model with the same dataset, or provide access to the model. In general. releasing code and data is often one good way to accomplish this, but reproducibility can also be provided via detailed instructions for how to replicate the results, access to a hosted model (e.g., in the case of a large language model), releasing of a model checkpoint, or other means that are appropriate to the research performed.
        \item While NeurIPS does not require releasing code, the conference does require all submissions to provide some reasonable avenue for reproducibility, which may depend on the nature of the contribution. For example
        \begin{enumerate}
            \item If the contribution is primarily a new algorithm, the paper should make it clear how to reproduce that algorithm.
            \item If the contribution is primarily a new model architecture, the paper should describe the architecture clearly and fully.
            \item If the contribution is a new model (e.g., a large language model), then there should either be a way to access this model for reproducing the results or a way to reproduce the model (e.g., with an open-source dataset or instructions for how to construct the dataset).
            \item We recognize that reproducibility may be tricky in some cases, in which case authors are welcome to describe the particular way they provide for reproducibility. In the case of closed-source models, it may be that access to the model is limited in some way (e.g., to registered users), but it should be possible for other researchers to have some path to reproducing or verifying the results.
        \end{enumerate}
    \end{itemize}

\item {\bf Open access to data and code}
    \item[] Question: Does the paper provide open access to the data and code, with sufficient instructions to faithfully reproduce the main experimental results, as described in supplemental material?
    \item[] Answer: \answerYes{}
    \item[] Justification: The supplementary material includes our training code for \ourMethod{}, the launch scripts encoding all training configurations and hyperparameters, and a README documenting the software environment and the exact commands required to run training. The training corpus (DeepScaleR) and all evaluation benchmarks are publicly available with citations in Appendix~\ref{app:datasets}, and all backbone models are released under open-source licenses. Baseline checkpoints are taken from the public releases of the original authors as documented in Appendix~\ref{app:baselines}. The full de-anonymized repository, including the evaluation pipeline and trained model checkpoints, will be released upon acceptance.
    \item[] Guidelines:
    \begin{itemize}
        \item The answer \answerNA{} means that paper does not include experiments requiring code.
        \item Please see the NeurIPS code and data submission guidelines (\url{https://neurips.cc/public/guides/CodeSubmissionPolicy}) for more details.
        \item While we encourage the release of code and data, we understand that this might not be possible, so \answerNo{} is an acceptable answer. Papers cannot be rejected simply for not including code, unless this is central to the contribution (e.g., for a new open-source benchmark).
        \item The instructions should contain the exact command and environment needed to run to reproduce the results. See the NeurIPS code and data submission guidelines (\url{https://neurips.cc/public/guides/CodeSubmissionPolicy}) for more details.
        \item The authors should provide instructions on data access and preparation, including how to access the raw data, preprocessed data, intermediate data, and generated data, etc.
        \item The authors should provide scripts to reproduce all experimental results for the new proposed method and baselines. If only a subset of experiments are reproducible, they should state which ones are omitted from the script and why.
        \item At submission time, to preserve anonymity, the authors should release anonymized versions (if applicable).
        \item Providing as much information as possible in supplemental material (appended to the paper) is recommended, but including URLs to data and code is permitted.
    \end{itemize}

\item {\bf Experimental setting/details}
    \item[] Question: Does the paper specify all the training and test details (e.g., data splits,
    hyperparameters, how they were chosen, type of optimizer) necessary to understand the results?
    \item[] Answer: \answerYes{}
    \item[] Justification: Section~\ref{sec:experiments:setup} summarises the experimental setup
    including backbone models, training dataset, evaluation benchmarks, and baselines.
    Appendix~\ref{app:training_params} provides the complete hyperparameter table.
    Appendix~\ref{app:eval_protocol} defines all evaluation metrics (Acc, Token, IPT,
    $\Delta$Acc, $\Delta$Token) with explicit formulas.
    \item[] Guidelines:
    \begin{itemize}
        \item The answer \answerNA{} means that the paper does not include experiments.
        \item The experimental setting should be presented in the core of the paper to a level of detail that is necessary to appreciate the results and make sense of them.
        \item The full details can be provided either with the code, in appendix, or as supplemental material.
    \end{itemize}

\item {\bf Experiment statistical significance}
\item[] Question: Does the paper report error bars suitably and correctly defined or other appropriate information about the statistical significance of the experiments?
\item[] Answer: \answerYes{}
\item[] Justification: For evaluation, we address statistical variance differently based on dataset size. For small-scale benchmarks (AIME 2024 and AMC 2023), we repeat evaluation 16 and 8 times respectively and report the mean to reduce random error caused by sampling variability. For larger benchmarks (e.g., MATH-500), the dataset size is sufficient to yield stable estimates, making repeated evaluation unnecessary. For training, we do not report error bars over multiple independent RL training runs due to the prohibitive computational cost (each run requires 8$\times$A800 GPUs for substantial wall-clock time), which is consistent with standard practice in the LLM RL literature.
\item[] Guidelines:
\begin{itemize}
    \item The answer \answerNA{} means that the paper does not include experiments.
    \item The authors should answer \answerYes{} if the results are accompanied by error bars, confidence intervals, or statistical significance tests, at least for the experiments that support the main claims of the paper.
    \item The factors of variability that the error bars are capturing should be clearly stated (for example, train/test split, initialization, random drawing of some parameter, or overall run with given experimental conditions).
    \item The method for calculating the error bars should be explained (closed form formula, call to a library function, bootstrap, etc.)
    \item The assumptions made should be given (e.g., Normally distributed errors).
    \item It should be clear whether the error bar is the standard deviation or the standard error of the mean.
    \item It is OK to report 1-sigma error bars, but one should state it. The authors should preferably report a 2-sigma error bar than state that they have a 96\% CI, if the hypothesis of Normality of errors is not verified.
    \item For asymmetric distributions, the authors should be careful not to show in tables or figures symmetric error bars that would yield results that are out of range (e.g., negative error rates).
    \item If error bars are reported in tables or plots, the authors should explain in the text how they were calculated and reference the corresponding figures or tables in the text.
\end{itemize}

\item {\bf Experiments compute resources}
    \item[] Question: For each experiment, does the paper provide sufficient information on the
    computer resources (type of compute workers, memory, time of execution) needed to
    reproduce the experiments?
    \item[] Answer: \answerYes{}
    \item[] Justification: Appendix~\ref{app:training_params} states that all experiments are run
    on 8$\times$A800 80\,GB GPUs using FSDP with CPU offloading, and specifies the training
    duration (300 steps). Wall-clock time per step is also reported in the ablation training
    curves (Appendix~\ref{app:ablation_curves}, Figure~\ref{fig:ablation_curves},
    Figure~\ref{fig:ablation_curves:app}).
    \item[] Guidelines:
    \begin{itemize}
        \item The answer \answerNA{} means that the paper does not include experiments.
        \item The paper should indicate the type of compute workers CPU or GPU, internal cluster, or cloud provider, including relevant memory and storage.
        \item The paper should provide the amount of compute required for each of the individual experimental runs as well as estimate the total compute.
        \item The paper should disclose whether the full research project required more compute than the experiments reported in the paper (e.g., preliminary or failed experiments that didn't make it into the paper).
    \end{itemize}

\item {\bf Code of ethics}
    \item[] Question: Does the research conducted in the paper conform, in every respect, with the NeurIPS Code of Ethics \url{https://neurips.cc/public/EthicsGuidelines}?
    \item[] Answer: \answerYes{}
    \item[] Justification: The paper studies training-time efficiency of reasoning models using publicly available datasets and open-source backbone models. No human subjects, sensitive data, or dual-use concerns are involved.
    \item[] Guidelines:
    \begin{itemize}
        \item The answer \answerNA{} means that the authors have not reviewed the NeurIPS Code of Ethics.
        \item If the authors answer \answerNo, they should explain the special circumstances that require a deviation from the Code of Ethics.
        \item The authors should make sure to preserve anonymity (e.g., if there is a special consideration due to laws or regulations in their jurisdiction).
    \end{itemize}

\item {\bf Broader impacts}
    \item[] Question: Does the paper discuss both potential positive societal impacts and negative societal impacts of the work performed?
    \item[] Answer: \answerYes{}
    \item[] Justification: The paper proposes a training-efficiency improvement for reasoning models, which has clear positive societal impacts: reducing inference cost lowers energy consumption and improves the accessibility of capable reasoning models for users with limited compute budgets. We do not identify a direct path to negative applications, as the work targets efficiency rather than capability expansion, releases no new pre-trained model or dataset, and does not involve sensitive domains.
    \item[] Guidelines:
    \begin{itemize}
        \item The answer \answerNA{} means that there is no societal impact of the work performed.
        \item If the authors answer \answerNA{} or \answerNo, they should explain why their work has no societal impact or why the paper does not address societal impact.
        \item Examples of negative societal impacts include potential malicious or unintended uses (e.g., disinformation, generating fake profiles, surveillance), fairness considerations (e.g., deployment of technologies that could make decisions that unfairly impact specific groups), privacy considerations, and security considerations.
        \item The conference expects that many papers will be foundational research and not tied to particular applications, let alone deployments. However, if there is a direct path to any negative applications, the authors should point it out. For example, it is legitimate to point out that an improvement in the quality of generative models could be used to generate Deepfakes for disinformation. On the other hand, it is not needed to point out that a generic algorithm for optimizing neural networks could enable people to train models that generate Deepfakes faster.
        \item The authors should consider possible harms that could arise when the technology is being used as intended and functioning correctly, harms that could arise when the technology is being used as intended but gives incorrect results, and harms following from (intentional or unintentional) misuse of the technology.
        \item If there are negative societal impacts, the authors could also discuss possible mitigation strategies (e.g., gated release of models, providing defenses in addition to attacks, mechanisms for monitoring misuse, mechanisms to monitor how a system learns from feedback over time, improving the efficiency and accessibility of ML).
    \end{itemize}

\item {\bf Safeguards}
    \item[] Question: Does the paper describe safeguards that have been put in place for responsible release of data or models that have a high risk for misuse (e.g., pre-trained language models, image generators, or scraped datasets)?
    \item[] Answer: \answerNA{}
    \item[] Justification: The paper proposes a training algorithm and does not release a new pre-trained model or dataset at submission time. The backbone models used (DeepSeek-R1-Distill, Qwen3) are governed by their respective open-source licenses.
    \item[] Guidelines:
    \begin{itemize}
        \item The answer \answerNA{} means that the paper poses no such risks.
        \item Released models that have a high risk for misuse or dual-use should be released with necessary safeguards to allow for controlled use of the model, for example by requiring that users adhere to usage guidelines or restrictions to access the model or implementing safety filters.
        \item Datasets that have been scraped from the Internet could pose safety risks. The authors should describe how they avoided releasing unsafe images.
        \item We recognize that providing effective safeguards is challenging, and many papers do not require this, but we encourage authors to take this into account and make a best faith effort.
    \end{itemize}

\item {\bf Licenses for existing assets}
    \item[] Question: Are the creators or original owners of assets (e.g., code, data, models), used in the paper, properly credited and are the license and terms of use explicitly mentioned and properly respected?
    \item[] Answer: \answerYes{}
    \item[] Justification: All backbone models (DeepSeek-R1-Distill-Qwen-1.5B/7B, Qwen3-8B), the training dataset (DeepScaleR), and all evaluation benchmarks (AIME24, AMC23, MATH-500, Minerva, OlympiadBench, LiveCodeBench, GPQA-Diamond, MMLU) are properly cited with their original papers. Appendix~\ref{app:datasets} provides detailed descriptions of all datasets.
    \item[] Guidelines:
    \begin{itemize}
        \item The answer \answerNA{} means that the paper does not use existing assets.
        \item The authors should cite the original paper that produced the code package or dataset.
        \item The authors should state which version of the asset is used and, if possible, include a URL.
        \item The name of the license (e.g., CC-BY 4.0) should be included for each asset.
        \item For scraped data from a particular source (e.g., website), the copyright and terms of service of that source should be provided.
        \item If assets are released, the license, copyright information, and terms of use in the package should be provided. For popular datasets, \url{paperswithcode.com/datasets} has curated licenses for some datasets. Their licensing guide can help determine the license of a dataset.
        \item For existing datasets that are re-packaged, both the original license and the license of the derived asset (if it has changed) should be provided.
        \item If this information is not available online, the authors are encouraged to reach out to the asset's creators.
    \end{itemize}

\item {\bf New assets}
    \item[] Question: Are new assets introduced in the paper well documented and is the documentation provided alongside the assets?
    \item[] Answer: \answerNA{}
    \item[] Justification: The paper does not introduce new datasets or benchmarks. The proposed method (ExpThink) is a training algorithm; code and model checkpoints will be released upon acceptance.
    \item[] Guidelines:
    \begin{itemize}
        \item The answer \answerNA{} means that the paper does not release new assets.
        \item Researchers should communicate the details of the dataset\slash code\slash model as part of their submissions via structured templates. This includes details about training, license, limitations, etc.
        \item The paper should discuss whether and how consent was obtained from people whose asset is used.
        \item At submission time, remember to anonymize your assets (if applicable). You can either create an anonymized URL or include an anonymized zip file.
    \end{itemize}

\item {\bf Crowdsourcing and research with human subjects}
    \item[] Question: For crowdsourcing experiments and research with human subjects, does the paper include the full text of instructions given to participants and screenshots, if applicable, as well as details about compensation (if any)?
    \item[] Answer: \answerNA{}
    \item[] Justification: The paper does not involve crowdsourcing or research with human subjects. All experiments are conducted entirely with automated model training and evaluation on existing benchmarks.
    \item[] Guidelines:
    \begin{itemize}
        \item The answer \answerNA{} means that the paper does not involve crowdsourcing nor research with human subjects.
        \item Including this information in the supplemental material is fine, but if the main contribution of the paper involves human subjects, then as much detail as possible should be included in the main paper.
        \item According to the NeurIPS Code of Ethics, workers involved in data collection, curation, or other labor should be paid at least the minimum wage in the country of the data collector.
    \end{itemize}

\item {\bf Institutional review board (IRB) approvals or equivalent for research with human subjects}
    \item[] Question: Does the paper describe potential risks incurred by study participants, whether such risks were disclosed to the subjects, and whether Institutional Review Board (IRB) approvals (or an equivalent approval/review based on the requirements of your country or institution) were obtained?
    \item[] Answer: \answerNA{}
    \item[] Justification: The paper does not involve human subjects research and therefore requires no IRB approval.
    \item[] Guidelines:
    \begin{itemize}
        \item The answer \answerNA{} means that the paper does not involve crowdsourcing nor research with human subjects.
        \item Depending on the country in which research is conducted, IRB approval (or equivalent) may be required for any human subjects research. If you obtained IRB approval, you should clearly state this in the paper.
        \item We recognize that the procedures for this may vary significantly between institutions and locations, and we expect authors to adhere to the NeurIPS Code of Ethics and the guidelines for their institution.
        \item For initial submissions, do not include any information that would break anonymity (if applicable), such as the institution conducting the review.
    \end{itemize}

\item {\bf Declaration of LLM usage}
    \item[] Question: Does the paper describe the usage of LLMs if it is an important, original, or non-standard component of the core methods in this research? Note that if the LLM is used only for writing, editing, or formatting purposes and does \emph{not} impact the core methodology, scientific rigor, or originality of the research, declaration is not required.
    \item[] Answer: \answerYes{}
    \item[] Justification: LLMs are the central subject of this research. The paper explicitly describes the three backbone LLMs used for training and evaluation: DeepSeek-R1-Distill-Qwen-1.5B, DeepSeek-R1-Distill-Qwen-7B, and Qwen3-8B. Their roles, training configurations, and evaluation results are fully documented in Section~\ref{sec:experiments} and Appendices~\ref{app:training_params}-\ref{app:datasets}.
    \item[] Guidelines:
    \begin{itemize}
        \item The answer \answerNA{} means that the core method development in this research does not involve LLMs as any important, original, or non-standard components.
        \item Please refer to our LLM policy in the NeurIPS handbook for what should or should not be described.
    \end{itemize}

\end{enumerate}

\end{document}